\documentclass[letterpaper,journal]{IEEEtran}
\usepackage{amsmath,amsfonts}
\usepackage{array}
\usepackage[caption=false,font=normalsize,labelfont=sf,textfont=sf]{subfig}
\usepackage{textcomp}
\usepackage{stfloats}
\usepackage{url}
\usepackage{verbatim}
\usepackage{graphicx}
\usepackage{cite}
\usepackage{booktabs,multirow,makecell,array,graphicx}
\usepackage{subfig}
\usepackage{multirow}
\usepackage{amsthm,amsmath,amssymb}
\usepackage{mathrsfs}
\usepackage{booktabs}
\usepackage[dvipsnames]{xcolor}
\usepackage{enumitem}
\usepackage[normalem]{ulem}
\usepackage{pifont}

\newcommand{\tabincell}[2]{\begin{tabular}{@{}#1@{}}#2\end{tabular}}

\newcounter{magicrownumbers}

\usepackage[ruled,vlined]{algorithm2e}
\definecolor{commentcolor}{RGB}{25, 25, 112}   %

\begin{document}

\title{MMGait: Benchmarking and Unifying Gait Recognition across Heterogeneous Modalities}

\author{Saihui~Hou,~\IEEEmembership{Member,~IEEE,}
    Chenye~Wang,
    Qingyuan~Cai,
    Aoqi~Li,
	Yongzhen~Huang,~\IEEEmembership{Senior Member,~IEEE} %
\thanks{Saihui Hou and Qingyuan Cai are with the School of Artificial Intelligence, Beijing Normal University, Beijing, China (e-mail: housaihui@bnu.edu.cn; caiqingyuan@mail.bnu.edu.cn).}
\thanks{Chenye Wang is with The Hong Kong University of Science and Technology, Hong Kong, China (e-mail: chenye.wang@connect.ust.hk). Aoqi Li is with JD.com, Beijing, China (e-mail: liaoqi16@jd.com). They contributed to this work while they were with the School of Artificial Intelligence, Beijing Normal University.}
\thanks{Yongzhen Huang is with the School of Artificial Intelligence, Beijing Normal University, Beijing, China, and also with Watrix Technology Limited Co. Ltd, Beijing, China (e-mail: huangyongzhen@bnu.edu.cn).}
}

\pagestyle{plain}

\maketitle
\thispagestyle{plain}

\begin{abstract}
Gait recognition is commonly studied using RGB videos or their derived
silhouettes and poses. Yet human walking produces heterogeneous photometric,
geometric, and motion cues that cannot be systematically examined with
RGB-centered benchmarks. We present \textbf{MMGait}, a large-scale
multi-sensor benchmark that brings visible, infrared, depth, LiDAR, and radar
observations into sequence-level correspondence. It provides diverse
modalities spanning appearance, contours, geometry, motion, and body
structure. Under a shared impostor-augmented protocol, we systematically
evaluate single-modal recognition, cross-modal recognition via directed
retrieval, and multi-modal recognition using task-specific experts. Across the
evaluated settings, modality rankings vary with probe conditions, cross-modal alignment remains
difficult, and fusion often provides complementary gains.
This analysis also exposes a scalability problem: individual modalities,
modality pairs, and fusion configurations are typically handled by separately
trained experts. We therefore formulate \textbf{Omni-Modal Gait Recognition},
which unifies single-modal, cross-modal, and multi-modal recognition within a
shared identity space. To instantiate this formulation, \textbf{OmniGait++}
uses modality-specific front ends followed
by a shared identity encoder to preserve modality-dependent cues while learning
comparable identity descriptors. An anchor-guided fusion module then
aggregates modality subsets of varying size without requiring frame-level
synchronization. A single jointly trained checkpoint operates across the three
recognition settings and accommodates modality subsets of different
compositions and cardinalities. Experiments show that OmniGait++ remains
competitive with task-specific experts in many shared settings and extends to
higher-cardinality fusion unavailable to fixed-pair models. These results
establish MMGait as a common testbed for heterogeneous gait sensing and
demonstrate the feasibility of unified recognition under varying modality
availability.
\end{abstract}


\section{Introduction}
\label{sec:intro}

Gait recognition identifies individuals from their walking patterns at a
distance and without requiring explicit cooperation, making it attractive for
non-contact biometric applications~\cite{sepas2022deep}.
Human walking is an inherently multi-faceted physical process whose
identity-related cues are conveyed through appearance, body shape,
three-dimensional structure, and motion. Yet most existing studies observe
this process through RGB videos or their derived modalities, particularly
silhouettes~\cite{chao2019gaitset,fan2023opengait} and
poses~\cite{liao2020model,fan2024skeletongait}.
Despite strong performance in controlled and unconstrained
environments~\cite{ye2024biggait,ye2025biggergait},
these modalities remain different abstractions of the same projected
visual source. This leaves a fundamental question: given
corresponding gait sequences, how discriminative is each modality, how well can
heterogeneous modalities be aligned, and to what extent can they complement
one another?

Existing multi-sensor benchmarks cannot adequately answer this question because
their sensing breadth remains limited. Earlier datasets provide heterogeneous
or non-RGB observations but contain only a few hundred identities and offer
sparse view coverage~\cite{tan2006efficient,hofmann2014tum}. Recent benchmarks
increase the population to more than one thousand identities, but largely
concentrate on RGB--LiDAR sensing~\cite{shen2023lidargait,han2024gait}.
Consequently,
photometric, geometric, and motion-related cues are rarely observed together
for the same identities, walking conditions, and viewpoints, making their
discriminability, redundancy, and complementarity difficult to study
systematically.

To overcome this bottleneck, we present \textbf{MMGait}, a large-scale
multi-sensor gait benchmark comprising \textbf{1{,}015 identities}, five sensing
streams, twelve gait modalities, ten walking directions,
and three walking conditions.
After quality
control, MMGait contains \textbf{482{,}327} gait-sequence instances spanning
visible and infrared (IR) appearance, body contours, dense depth, 3D point clouds,
projected geometry, event-based motion, and 2D/3D pose. By bringing these
photometric, geometric, and motion-related
observations together with sequence-level correspondence, MMGait provides a
common testbed for examining how heterogeneous gait cues differ, align, and
complement one another.

\begin{figure*}[t]
	\centering
	\includegraphics[width=0.99\linewidth]{figures/overview.pdf}
	\caption{Overview of MMGait. Five sensing streams provide twelve raw or
		derived gait modalities across ten views and three walking
		conditions.}
	\label{fig:mmgait_overview}
\vspace{-2ex}
\end{figure*}

Using this common data foundation, we establish a unified, impostor-augmented
protocol to evaluate heterogeneous gait sensing in three settings:
single-modal recognition for standalone discriminability, cross-modal
recognition via directed retrieval for identity alignment, and multi-modal
recognition via cue fusion for complementarity. We benchmark all twelve
modalities individually and, given the rapidly growing configuration space, representative directed
cross-modal retrieval tasks and two-modality fusion configurations
using task-specific experts. Results show that robustness varies across
modalities and probe conditions, cross-modal alignment remains challenging and pair-dependent, and
fusion often provides substantial gains over individual modalities. Together,
these findings show that a gait modality's utility reflects its
standalone discriminability, cross-modal compatibility, and complementarity
with other sensing cues.

This systematic evaluation also exposes a deeper scalability problem: each
single modality, cross-modal pair, and fusion subset is typically handled by a
separately trained expert. The number of such models grows rapidly as the
supported modality set expands, making exhaustive development and deployment
impractical. We therefore formulate \textbf{Omni-Modal Gait Recognition}, which
treats the input as a set of available modalities rather than a fixed modality
configuration and
learns one shared identity space for single-modal recognition, cross-modal
recognition, and multi-modal recognition. Its central challenge is to preserve
the distinctive physical information of each modality while maintaining identity
comparability as the number and composition of the available inputs change.

To address this challenge, we propose \textbf{OmniGait++}. Unlike conventional
methods designed for a predefined modality or modality combination, OmniGait++
supports different modality subsets within a unified model. Its
modality-specific encoders preserve distinctive input characteristics, while
shared representation learning maps heterogeneous observations into a common
identity space. When multiple modalities are available, an adaptive fusion
module integrates their features and recalibrates their contributions according
to the modality combination. We evaluate a single jointly trained model across
9 single-modal settings, 12 directed cross-modal retrieval tasks, and 19
variable-cardinality fusion configurations.
Single-modal recognition, cross-modal recognition, and multi-modal recognition
thus become different operating settings of one shared framework rather than
separate models.

In summary, our contributions are threefold:
\begin{itemize}
	\item We construct \textbf{MMGait}, a large-scale benchmark that contains
	five heterogeneous sensing streams and provides twelve gait modalities
	for 1{,}015 identities. Its dense cross-view coverage and sequence-level
	correspondence enable physically different gait cues to be studied within
	one common experimental basis.
	\item We establish a common, impostor-augmented evaluation across
	single-modal recognition, cross-modal recognition, and multi-modal
	recognition.
	Beyond ranking individual modalities, the resulting study characterizes
	their robustness across probe conditions, cross-modal identity compatibility,
	and complementary value.
	\item We formulate \textbf{Omni-Modal Gait Recognition} and propose
	\textbf{OmniGait++}, whose shared identity learning and adaptive
	variable-cardinality fusion enable a single jointly trained model to support
	single-modal recognition, cross-modal recognition, and multi-modal
	recognition across different modality subsets.
\end{itemize}

This article substantially extends our preliminary conference
work~\cite{wan2026mmgait} along the data, evaluation, and modeling dimensions.
First, we re-audit the original 725-participant collection for multi-sensor
recording completeness, retain recordings from 685 participants, and add data
from 330 newly recorded participants, producing a 1{,}015-identity benchmark
with improved data completeness. Second, we introduce an
impostor-augmented evaluation protocol and
re-run all single-modal, cross-modal, and multi-modal experiments on the
expanded benchmark rather than reusing conference results. Third, we formulate
Omni-Modal Gait Recognition and implement it by extending the pairwise
\textbf{OmniGait} model to \textbf{OmniGait++} with adaptive
variable-cardinality fusion. We also establish a 19-configuration registry for
evaluating variable-cardinality fusion. Together, these
additions provide a substantially expanded benchmark, a more challenging
evaluation protocol, and a unified model for variable-cardinality fusion.

\section{Related Work}
\label{sec:related}

\subsection{Gait Recognition with Diverse Modalities}

Binary silhouettes remain the dominant gait modality because they
suppress appearance and background details while preserving body shape and
motion. GaitSet treats a sequence as an unordered
set~\cite{chao2019gaitset}, whereas GaitPart learns localized temporal patterns
from horizontal body regions~\cite{fan2020gaitpart}. GaitGL subsequently
combines global and local representations~\cite{lin2022gaitgl}. More recent
studies revisit general-purpose architectures at larger scales: GaitBase
establishes a strong convolutional baseline, and DeepGaitV2 examines the effect
of network depth~\cite{fan2023opengait,fan2023exploring}.
Other methods improve robustness through causal analysis, motion modeling, or
clothing-related representation learning~\cite{xiong2024causality,ma2023dynamic,ma2024learning,huang2025learning,yang2025bridging}.
Despite this progress,
silhouettes depend on foreground segmentation. They also discard texture,
depth, and other physical measurements.

Pose modalities provide a complementary description based on body joints
and their temporal dynamics. PoseGait extracts identity cues from estimated 3D
joint trajectories~\cite{liao2020model}, while GaitGraph and GaitGraph2 model
skeleton sequences as spatio-temporal
graphs~\cite{teepe2021gaitgraph,teepe2022towards}. GPGait introduces human-prior-guided
pose modeling~\cite{fu2023gpgait}, and subsequent studies improve pose
estimation or temporal representation~\cite{meng2025fastposegait,guo2025gaitcontour}.
SkeletonGait instead renders poses as heatmaps, retaining spatial structure for
convolutional processing~\cite{fan2024skeletongait}. Pose is compact and less
dependent on texture, but recognition remains sensitive to estimation errors
under occlusion, low resolution, or unusual viewpoints.

Recent work broadens gait recognition toward appearance, spectral, temporal,
geometric, and radio-frequency information. RGB retains texture and color.
BigGait and BiggerGait use large vision models to extract gait-related features
from RGB videos~\cite{ye2024biggait,ye2025biggergait}. Infrared imaging supplies
spectrum-dependent appearance under weak visible
illumination~\cite{tan2006efficient}. Event-based approaches emphasize temporal changes:
EdinoGait studies event representation learning~\cite{chen2025edinogait}, while
EventGait separates high-frequency motion and shape
cues~\cite{xu2026eventgait}. Depth cameras provide dense geometry with reduced
dependence on texture~\cite{haque2016recurrent}. PointGait processes 3D
point-cloud sequences~\cite{wang2023pointgait}, whereas LidarGait and
LidarGait++ develop LiDAR-specific representations~\cite{shen2023lidargait,shen2025lidargait++}.
Radar further captures motion responses relevant to gait identification under
weak visual conditions~\cite{meng2020gait}. These sensing mechanisms expose
complementary physical properties, yet they are generally studied on separate
datasets with dedicated models. Controlled comparison and joint modeling
across heterogeneous sensing modalities therefore remain difficult.

\subsection{Cross-Modal, Multi-Modal, and Unified Gait Recognition}

Cross-modal gait recognition matches a probe in one modality against a gallery
represented in another and is typically evaluated through directed retrieval.
Existing gait methods mainly study this task between cameras and LiDAR.
CrossGait learns modality-shared prototypes and
feature adapters~\cite{wang2024cross}, whereas CL-Gait uses contrastive
pre-training with paired silhouettes and synthetic point
clouds~\cite{guo2024camera}.
TCFDNet further employs text-guided feature disentanglement to separate shared
identity information from modality-specific
characteristics~\cite{lu2026textguided}. These methods improve 2D--3D alignment, and some also
preserve within-modality recognition. Their cross-modal components,
nevertheless, remain optimized for a predefined modality pair.

Multi-modal gait recognition instead exploits complementary observations that
are simultaneously available. MMGaitFormer and TriGait combine silhouette--pose
cues through multi-branch architectures~\cite{cui2023multi,sun2024trigait}.
MultiGait++ investigates shared and distinctive information in silhouettes,
human parsing, and optical flow~\cite{jin2025exploring}. Beyond RGB-derived
modalities, EMGaitNet integrates camera--LiDAR observations using semantic
guidance~\cite{lu2026walking}. These
methods establish the value of complementary observations, but most assume a
fixed pair or a small predetermined modality set. UGaitNet relaxes this
assumption by accepting a variable number of grayscale, optical flow, depth, or
silhouette inputs~\cite{marin2021ugaitnet}. Its inputs, however, remain
primarily camera-centric.

Beyond gait recognition, person ReID provides related progress toward a common
identity space across heterogeneous inputs. All-in-One learns consistent
representations from RGB, infrared, sketch, and text through a modality-aware
shared model~\cite{li2024all}. ReID5o introduces a five-modality benchmark and
expert routing for variable query combinations~\cite{zuo2025reid5o}. These
studies demonstrate the feasibility of unified identity modeling, but they
primarily concern appearance-based or semantic inputs. Multi-sensor gait introduces
further physical heterogeneity because its modalities encode spectrum-dependent
appearance, geometry at different densities, sparse motion responses, or body
structure. Existing methods thus cover only part of the target task space.
Omni-Modal Gait Recognition instead unifies single-modal recognition,
cross-modal recognition, and multi-modal recognition over variable-cardinality
modality subsets within a shared identity space.

\subsection{Gait Recognition Benchmarks}

Controlled RGB benchmarks established the standard settings for cross-view and
cross-condition gait recognition~\cite{li2024aerialgait,wang2025ra}. CASIA-B includes normal walking, bag
carrying, and clothing change across multiple views~\cite{yu2006framework},
whereas OU-MVLP substantially increases the population scale for cross-view
evaluation~\cite{takemura2018multi}. CCPG emphasizes clothing
variation~\cite{li2023depth}. CCGR provides larger-scale cross-view and cross-condition
observations~\cite{zou2024cross}. Gait3D and GREW extend gait recognition to
unconstrained scenes, trajectories, and pedestrian
appearances~\cite{zheng2022gait,zhu2021gait}. These benchmarks have driven substantial
progress, but their commonly used modalities are predominantly derived
from RGB videos.

Non-RGB and multi-sensor benchmarks broaden the sensing scope. CASIA-C provides
thermal-infrared sequences captured at night~\cite{tan2006efficient}, while
TUM-GAID combines RGB, depth, and audio observations~\cite{hofmann2014tum}.
SUSTech1K provides synchronized camera--LiDAR data for point-cloud gait
recognition~\cite{shen2023lidargait}. FreeGait extends camera--LiDAR acquisition
to unconstrained outdoor trajectories~\cite{han2024gait}, whereas LRGait
emphasizes synchronized long-range and cross-distance
sensing~\cite{lu2026walking}. These datasets demonstrate the value of cross-spectrum
or 3D observations, but each covers only a limited portion of the sensor
combinations encountered in heterogeneous perception systems. MMGait instead
combines broader sensing diversity with dense multi-view coverage and
sequence-level correspondence under a unified evaluation framework, enabling
controlled comparison of physically different gait modalities.

\section{MMGait Benchmark}
\label{sec:mmgait_benchmark}

MMGait is a large-scale multi-sensor gait benchmark containing 1{,}015
identities and 482{,}327 released gait sequences. As illustrated in
Fig.~\ref{fig:mmgait_overview}, the benchmark provides five sensing streams:
RGB, depth, IR, LiDAR, and 4D radar. These streams are processed into twelve
gait modalities spanning visible and IR appearance, body contours,
2D/3D pose, dense depth, 3D point clouds, projected geometry, and event-based
motion. Each participant is recorded from ten walking directions under
normal walking, backpack carrying, and clothing change. MMGait therefore
jointly provides population scale, multi-view
coverage, and sensing diversity for studying single-modal recognition,
cross-modal recognition, multi-modal recognition, and unified Omni-Modal Gait
Recognition.

\subsection{Multi-Sensor Data Acquisition}

Table~\ref{tab:mmgait_sensors} summarizes the sensor specifications and
resulting gait modalities. The RGB and depth streams are captured using
an Orbbec Gemini 2 XL RGB-D camera module~\cite{Orbbec}. The RGB stream records
visible appearance, whereas the infrared-enhanced stereo depth stream captures
dense body geometry with reduced dependence on texture. IR videos are
recorded by an independent industrial camera operating at a 940-nm narrowband
wavelength. The 940-nm wavelength was selected to reduce interference from the
LiDAR laser while providing spectrum-dependent appearance under weak visible
illumination.
The LiDAR stream is acquired using an Ouster OS0-7
scanner~\cite{Ouster}, which records 3D point clouds. The GPAL Ares-R7861 4D
radar~\cite{Radar} provides sparse point clouds that encode complementary
spatial and motion responses.

\begin{table}[t]
	\centering
	\caption{Specifications of the five sensing streams and their twelve raw or derived modalities in MMGait.}
	\label{tab:mmgait_sensors}
	\setlength{\tabcolsep}{8pt}
	\renewcommand{\arraystretch}{1.0}
	\resizebox{0.95\columnwidth}{!}{%
	\begin{tabular}{@{}lccc@{}}
		\toprule
		\textbf{Sensor} & \textbf{Range} & \textbf{Resolution} &
		\textbf{Modalities} \\
		\midrule
		RGB & 0.4--10~m & 1280$\times$800 &
		\tabincell{c}{RGB, RGB Silhouette,\\2D/3D Pose, Event} \\
		\midrule
		IR & 0.4--10~m & 1280$\times$700 &
		\tabincell{c}{IR, IR Silhouette} \\
		\midrule
		Depth & 0.4--10~m & 1280$\times$800 & Depth \\
		\midrule
		LiDAR & 0.5--100~m & 128 beams &
		\tabincell{c}{Point Cloud,\\Projected Depth} \\
		\midrule
		4D Radar & 0.2--170~m & -- &
		\tabincell{c}{Point Cloud,\\Projected Depth} \\
		\bottomrule
	\end{tabular}}
\vspace{-2ex}
\end{table}

\begin{figure}[t]
	\centering
	\includegraphics[width=0.95\linewidth]{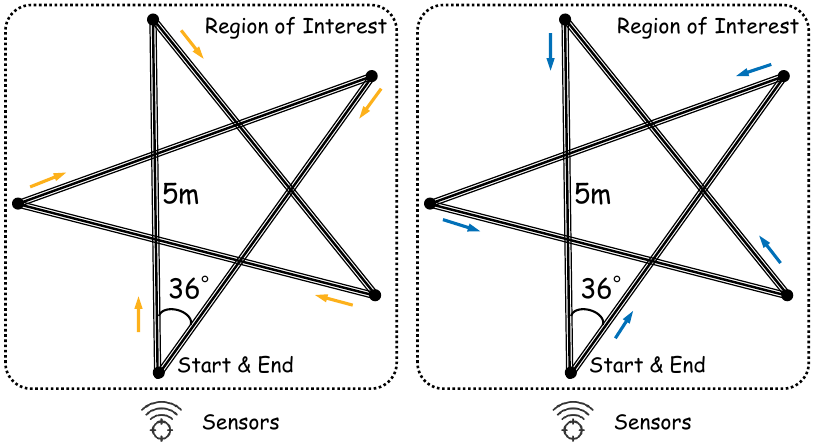}
	\caption{Five-pointed-star route used for MMGait acquisition. The two panels
		show opposite traversals of five segments, together yielding ten
		walking directions at approximately $36^{\circ}$ intervals relative to the
		fixed sensing system.}
	\label{fig:mmgait_route}
\vspace{-2ex}
\end{figure}

Using this acquisition system, participants walk along the five-pointed-star
route shown in Fig.~\ref{fig:mmgait_route}. The route produces ten walking
directions distributed over a full circle at intervals of approximately
$36^{\circ}$. Each walking direction is assigned a view label according to its
orientation relative to the fixed sensing system. Each participant is recorded
four times while walking the full route under three walking conditions: twice
during normal walking, once while carrying a backpack, and once after a
clothing change. Each full-route recording is divided into ten
direction-specific gait sequences following the predefined route segments.
The resulting sequences have four sequence types---NM-01 and NM-02 for normal
walking, BG-01 for backpack carrying, and CL-01 for clothing change---and each
sequence inherits the view label of its walking direction.
Participants use their own backpacks and
independently change both upper and lower garments, introducing natural
intra-condition diversity instead of using a single standardized accessory or
outfit.

During each recording, the five sensing streams observe the same participant
under a common walking condition. RGB and depth originate from the same RGB-D
module, whereas the IR, LiDAR, and radar streams use separate acquisition
interfaces without a common hardware trigger or time base. MMGait therefore
retains the native sequences and associates them at the sequence level using
the identity label, sequence type, and view label, without imposing potentially
inaccurate frame-level alignment. This design supports cross- and multi-modal
recognition without frame-level synchronization.

\subsection{Data Processing and Statistics}

Because the sensing mechanisms and data structures differ substantially, we
construct a stream-specific processing pipeline for each sensing stream. The
pipelines transform the raw recordings into modality-specific gait
sequences while preserving the physical characteristics of individual
modalities.

\textbf{RGB camera.}
We first apply pedestrian tracking~\cite{zhang2022bytetrack} to obtain human
bounding boxes from RGB videos. The resulting boxes are used to crop the RGB
appearance sequences. Within each box, a segmentation model extracts the
corresponding binary silhouette~\cite{liu2021paddleseg}. Beyond appearance and
silhouette extraction, the RGB videos are also used to derive pose and event
data. Pose estimators generate the 2D and 3D pose
sequences~\cite{wang2020deep,cai2024disentangled}, while
V2E~\cite{hu2021v2e} converts the RGB videos into event sequences.

\textbf{IR camera.}
Pedestrian tracking is performed independently on the IR videos to obtain
human bounding boxes. The boxes are used to crop the IR appearance sequences,
while pedestrian segmentation within each box produces the corresponding binary
silhouettes.

\textbf{Depth camera.}
Because RGB and depth are captured by the same RGB-D module, we transform the
RGB-derived bounding boxes into depth-image coordinates and use them to crop
the corresponding human depth regions.

\textbf{LiDAR scanner.}
We first crop the LiDAR scans to the predefined region of interest covering
the walking route. Ground points are removed using
Patchwork++~\cite{lee2022patchwork++}, followed by density-based
filtering~\cite{ester1996density} to suppress scattered noise and irrelevant
objects.
The filtered points form LiDAR point-cloud sequences and are further converted
into 2D projected-depth maps following established LiDAR gait
processing~\cite{shen2023lidargait}.

\textbf{4D radar.}
For the radar stream, we use the corresponding walking-area region of interest
to discard off-route detections. The retained sparse detections form radar
point-cloud sequences, from which 2D projected-depth maps are generated using
the LiDAR projection procedure described above.

\begin{figure}[!t]
	\centering
	\includegraphics[width=0.95\linewidth]{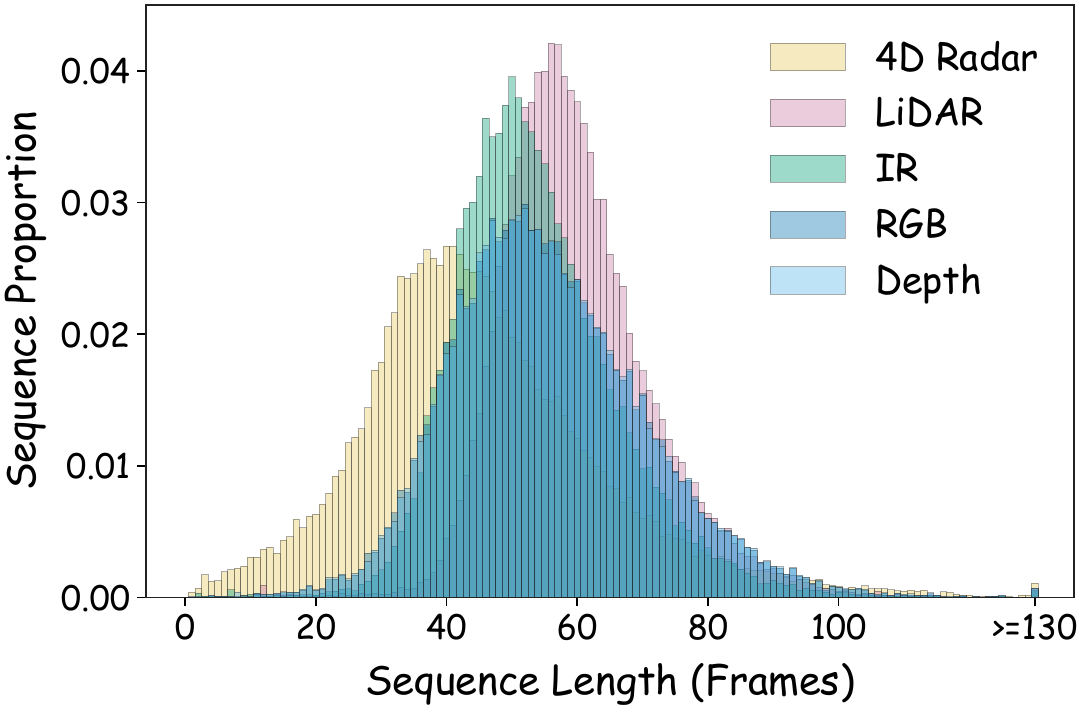}
	\caption{Sequence-length distributions of MMGait across sensing streams.}
	\label{fig:mmgait_seq_length}
\vspace{-2ex}
\end{figure}

After the stream-specific processing described above, the resulting
sequence-length distributions vary across sensing streams, as shown in
Fig.~\ref{fig:mmgait_seq_length}. This variation reflects differences in
acquisition rates and independently determined temporal boundaries.
Consistent with the sequence-level association described in the preceding
subsection, the released sequences retain their original frame counts and
are not temporally resampled to a uniform length. Beyond these
within-dataset statistics, Table~\ref{tab:gait_dataset_comparison} compares
MMGait with representative gait benchmarks.
Notably, MMGait complements existing benchmarks with controlled ten-direction
observations from five heterogeneous sensing streams under a common identity
and condition structure.

\begin{table*}[t]
	\centering
	\caption{Comparison of MMGait with representative gait-recognition benchmarks.}
	\label{tab:gait_dataset_comparison}
	\setlength{\tabcolsep}{6pt}
	\renewcommand{\arraystretch}{1.0}
	\resizebox{0.95\textwidth}{!}{%
	\begin{tabular}{@{}lccclll@{}}
		\toprule
		\textbf{Dataset} & \textbf{Year} & \textbf{Identities} &
		\textbf{Views} & \textbf{Sequences} & \textbf{Sensing Streams} &
		\textbf{Highlights} \\
		\midrule
		CASIA-B~\cite{yu2006framework} & 2006 & 124 & 11 & 13{,}640 & RGB & Cross-View and Covariates \\
		OU-MVLP~\cite{takemura2018multi} & 2018 & 10{,}307 & 14 & 288{,}596 & RGB & Cross-View \\
		GREW~\cite{zhu2021gait} & 2021 & 26{,}345 & 882 & 128{,}671 & RGB & In-the-Wild \\
		Gait3D~\cite{zheng2022gait} & 2022 & 4{,}000 & 39 & 25{,}309 & RGB & In-the-Wild \\
		CCPG~\cite{li2023depth} & 2023 & 200 & 10 & 16{,}566 & RGB & Clothing Change \\
		CCGR~\cite{zou2024cross} & 2024 & 970 & 33 & 1{,}580{,}617 & RGB & Cross-Condition \\
		\midrule
		CASIA-C~\cite{tan2006efficient} & 2006 & 153 & 1 & 1{,}530 & Thermal Infrared & Nighttime Thermal \\
		TUM-GAID~\cite{hofmann2014tum} & 2012 & 305 & 1 & 3{,}370 & RGB, Depth, Audio & RGB--Depth--Audio Covariates \\
		SUSTech1K~\cite{shen2023lidargait} & 2023 & 1{,}050 & 12 & 25{,}239 & RGB, LiDAR & Camera--LiDAR \\
		FreeGait~\cite{han2024gait} & 2024 & 1{,}195 & 1 & 11{,}950 & RGB, LiDAR & Free-Trajectory \\
		\textbf{MMGait} & \textbf{2026} & \textbf{1{,}015} & \textbf{10} &
		\textbf{482{,}327} &
		\textbf{RGB, IR, Depth, LiDAR, 4D Radar} &
		\textbf{Five-Stream Heterogeneous Sensing} \\
		\bottomrule
	\end{tabular}}
\vspace{-2ex}
\end{table*}

\subsection{Unified Evaluation Protocols}
\label{sec:mmgait_protocols}

We use an identity-disjoint split, with 300 identities for training and the
remaining 715 for testing. To isolate the effect of gallery composition, we
define two protocols that share the same probes and positive templates but
differ in their non-target samples. Protocol V1 evaluates matching against a
compact, view-specific normal-walking template gallery, whereas Protocol V2
evaluates robustness to a substantially larger impostor gallery.

Both protocols apply to single-modal recognition, cross-modal recognition, and
multi-modal recognition. Cross-modal recognition is evaluated as directed
probe-to-gallery retrieval. We denote the evaluated modality
configuration---a single modality, an ordered probe--gallery modality pair, or
a modality subset, respectively---by $c$. Let $\mathcal{X}_{g}^{(c)}$ denote
the available gallery-side test instances under $c$. Each
$x_i\in\mathcal{X}_{g}^{(c)}$ has identity label $y_i$, view label $v_i$, and
sequence type $s_i$.

BG-01 and CL-01 instances in the probe-side configuration serve as queries. For
a query with identity label $y$ and a designated template view $v_t$, its
available NM-01 and NM-02 gallery-side instances with identity label $y$ at
$v_t$ are positive. Separate $10\times10$ probe-view-by-template-view matrices
are constructed for BG and CL. Rank-1 accuracy and average precision (AP) are
computed for each query against the protocol-specific gallery and averaged
within each matrix cell. To focus on cross-view performance, the final Rank-1
accuracy and mean average precision (mAP) are obtained by uniformly averaging
the off-diagonal cells.

\subsubsection{Protocol V1: Template-Only Gallery}

Let $\mathcal{S}_{\mathrm{NM}}=\{\text{NM-01},\text{NM-02}\}$ denote the
normal-walking sequence types. For a positive-template view $v_t$, the gallery
contains the available templates of all test identities at that view:
\begin{equation}
	\mathcal{G}_{\mathrm{V1}}^{(c)}(v_t)=
	\{x_i\in\mathcal{X}_{g}^{(c)}\mid
	v_i=v_t,\;s_i\in\mathcal{S}_{\mathrm{NM}}\}.
\end{equation}

\subsubsection{Protocol V2: Impostor-Augmented Gallery}

For a query with identity label $y$ and a positive-template view $v_t$, Protocol
V2 augments the V1 gallery with every available gallery-side instance from
non-target identities, regardless of view or sequence type:
\begin{equation}
	\mathcal{G}_{\mathrm{V2}}^{(c)}(y,v_t)=
	\mathcal{G}_{\mathrm{V1}}^{(c)}(v_t)
	\cup\{x_i\in\mathcal{X}_{g}^{(c)}\mid y_i\neq y\}.
\end{equation}
All other instances with identity label $y$, including normal-walking templates
at other views, are excluded. Thus, the positive set remains identical to that
of Protocol V1.

\subsection{Ethics, Privacy, and Responsible Use}
\label{sec:mmgait_ethics}

\textbf{Consent and Pseudonymization.}
Participation was voluntary, and all participants provided written informed
consent before recording. The consent procedure explicitly covered the
collection of multi-sensor gait data, its use in academic research and
publications, and its release to qualified researchers. Direct personal
identifiers and demographic attributes are not included in the released
benchmark. Dataset annotations identify each participant only by a
pseudonymous participant ID. No mapping between these IDs and participants'
real-world identities is retained.

\textbf{Controlled Access and Permitted Use.}
MMGait is distributed through an application process rather than
unrestricted public download. Applicants must identify their institutional
affiliation and intended research purpose and agree to the dataset license
before access is granted. The license restricts the benchmark to research use
and prohibits redistribution. These conditions apply to both the raw sensor
observations and all derived modality data.

\textbf{Residual Risks and Participant Rights.}
Pseudonymization cannot eliminate privacy risks because sensor observations
and derived modality data may retain biometric information. Users are
therefore required to handle MMGait as sensitive human-subject data and comply
with the access conditions throughout their research. Participants may contact
the authors to request the removal of their data from subsequent releases. The
authors also provide a contact channel for questions about data access and for
reporting potential misuse.

\section{Systematic Evaluation of Gait Modalities}
\label{sec:modality_evaluation}

Before introducing a unified omni-modal model, we systematically characterize
the gait modalities provided by MMGait through three complementary
questions. First, how much standalone identity information does each
modality preserve under different probe conditions and gallery
definitions? Second, to what extent can identity descriptors be compared across
heterogeneous modalities? Third, when multiple modalities are jointly
available, do they provide complementary information beyond what each modality
provides individually? We address these questions using representative
task-specific baselines.
All evaluations follow the official identity-disjoint split, gallery
construction, and metrics defined in Section~\ref{sec:mmgait_protocols}. For
single-modal recognition, we report both Protocol V1 and Protocol V2 to
characterize sensitivity to the impostor-augmented gallery, whereas the
cross-modal and multi-modal evaluations focus on Protocol V2.

\subsection{Single-Modal Recognition}

\subsubsection{Single-Modal Baseline}

Single-modal baselines are selected according to the data structure of each
modality. For the established RGB-silhouette and pose modalities,
we compare multiple specialized methods. The pose comparison includes
coordinate-based methods for both 2D and 3D pose, together with the
heatmap-based SkeletonGait~\cite{fan2024skeletongait} for 2D pose. For
systematic coverage, DeepGaitV2-P3D~\cite{fan2023exploring} serves as the
common baseline for image-like inputs, while
LidarGait++~\cite{shen2025lidargait++} processes native LiDAR and radar point
clouds. Each method--modality combination is trained independently.
Table~\ref{tab:eval_single_modal} provides method-level comparisons for RGB
silhouettes and poses, together with task-appropriate baselines for all twelve
MMGait modalities.

\begin{table*}[t]
	\centering
	\caption{Single-modal recognition on MMGait under Protocols V1 and V2.
		Results are reported as Rank-1 accuracy and mAP (\%). Bold values denote
		the best result when multiple methods are compared for the same
		modality.}
	\label{tab:eval_single_modal}
	\small
	\setlength{\tabcolsep}{6pt}
	\renewcommand{\arraystretch}{1.0}
	\resizebox{0.80\textwidth}{!}{%
	\begin{tabular}{@{}lllcccccccc@{}}
		\toprule
		\multirow{3}{*}{\textbf{Sensing Streams}} & \multirow{3}{*}{\textbf{Modality}} &
		\multirow{3}{*}{\textbf{Method}} & \multicolumn{4}{c}{\textbf{Protocol V1}} &
		\multicolumn{4}{c}{\textbf{Protocol V2}} \\
		\cmidrule(lr){4-7}\cmidrule(lr){8-11}
		& & & \multicolumn{2}{c}{\textbf{BG}} & \multicolumn{2}{c}{\textbf{CL}} &
		\multicolumn{2}{c}{\textbf{BG}} & \multicolumn{2}{c}{\textbf{CL}} \\
		\cmidrule(lr){4-5}\cmidrule(lr){6-7}\cmidrule(lr){8-9}\cmidrule(lr){10-11}
		& & & R1 & mAP & R1 & mAP & R1 & mAP & R1 & mAP \\
		\midrule
		\multirow{15}{*}{RGB}
		& \multirow{5}{*}{Silhouette} & GaitSet~\cite{chao2019gaitset} & 92.7 & 92.3 & 55.7 & 60.0 & 79.4 & 77.9 & 29.9 & 31.9 \\
		& & GaitPart~\cite{fan2020gaitpart} & 95.4 & 95.1 & 59.9 & 64.1 & 83.5 & 82.5 & 33.1 & 35.3 \\
		& & GaitGL~\cite{lin2022gaitgl} & 94.3 & 93.4 & 57.4 & 61.3 & 73.2 & 73.6 & 24.2 & 27.7 \\
		& & GaitBase~\cite{fan2023opengait} & 98.0 & 97.9 & 72.3 & 75.7 & 92.0 & 91.6 & 45.0 & 47.4 \\
		& & DeepGaitV2-P3D~\cite{fan2023exploring} & \textbf{98.9} & \textbf{98.8} & \textbf{73.7} & \textbf{77.2} & \textbf{94.0} & \textbf{93.7} & \textbf{47.0} & \textbf{49.5} \\
		\cmidrule(lr){2-11}
		& \multirow{6}{*}{2D Pose} & GaitGraph~\cite{teepe2021gaitgraph} & 41.7 & 45.9 & 26.0 & 31.6 & 15.6 & 17.6 & 7.3 & 9.2 \\
		& & GaitGraph2~\cite{teepe2022towards} & 12.7 & 15.0 & 7.9 & 10.5 & 3.8 & 4.5 & 1.6 & 2.1 \\
		& & GaitTR~\cite{zhang2023spatial} & 66.1 & 68.1 & 47.2 & 51.5 & 37.3 & 38.4 & 21.4 & 23.6 \\
		& & GPGait~\cite{fu2023gpgait} & 74.2 & 74.7 & 48.6 & 52.2 & 42.8 & 44.8 & 21.6 & 24.2 \\
		& & GPGait++~\cite{meng2025fastposegait} & 79.5 & 80.0 & \textbf{59.8} & \textbf{62.7} & 53.4 & 54.3 & \textbf{33.7} & \textbf{35.7} \\
		& & SkeletonGait~\cite{fan2024skeletongait} & \textbf{80.9} & \textbf{80.9} & 52.8 & 56.0 & \textbf{56.1} & \textbf{56.3} & 27.6 & 29.4 \\
		\cmidrule(lr){2-11}
		& \multirow{2}{*}{3D Pose} & GPGait~\cite{fu2023gpgait} & 35.3 & 38.8 & 22.5 & 26.8 & 7.2 & 10.3 & 3.5 & 5.6 \\
		& & GPGait++~\cite{meng2025fastposegait} & \textbf{38.2} & \textbf{41.6} & \textbf{29.6} & \textbf{33.6} & \textbf{9.9} & \textbf{12.9} & \textbf{6.8} & \textbf{9.4} \\
		\cmidrule(lr){2-11}
		& RGB & DeepGaitV2-P3D~\cite{fan2023exploring} & 99.6 & 99.7 & 58.3 & 66.1 & 99.2 & 99.2 & 38.5 & 42.0 \\
		& Event & DeepGaitV2-P3D~\cite{fan2023exploring} & 64.5 & 67.6 & 19.7 & 24.8 & 28.7 & 31.9 & 5.3 & 7.0 \\
		\midrule
		\multirow{2}{*}{IR} & IR & DeepGaitV2-P3D~\cite{fan2023exploring} & 99.2 & 99.2 & 76.1 & 81.2 & 97.2 & 97.2 & 54.8 & 58.3 \\
		& Silhouette & DeepGaitV2-P3D~\cite{fan2023exploring} & 96.2 & 96.0 & 55.5 & 60.4 & 85.4 & 84.6 & 25.5 & 28.3 \\
		\midrule
		Depth & Depth & DeepGaitV2-P3D~\cite{fan2023exploring} & 93.2 & 93.1 & 73.7 & 76.7 & 72.6 & 73.0 & 42.9 & 45.4 \\
		\midrule
		\multirow{2}{*}{LiDAR} & Projected Depth & DeepGaitV2-P3D~\cite{fan2023exploring} & 94.9 & 95.5 & 79.4 & 82.5 & 83.9 & 84.6 & 55.6 & 58.0 \\
		& Point Cloud & LidarGait++~\cite{shen2025lidargait++} & 93.9 & 94.5 & 75.6 & 80.0 & 81.2 & 81.6 & 52.3 & 54.7 \\
		\midrule
		\multirow{2}{*}{4D Radar} & Projected Depth & DeepGaitV2-P3D~\cite{fan2023exploring} & 12.9 & 18.4 & 11.2 & 17.0 & 1.9 & 3.0 & 1.2 & 2.4 \\
		& Point Cloud & LidarGait++~\cite{shen2025lidargait++} & 26.8 & 34.6 & 23.2 & 31.6 & 5.1 & 7.5 & 3.6 & 5.9 \\
		\bottomrule
	\end{tabular}}
\vspace{-2ex}
\end{table*}

\subsubsection{Sensitivity to Gallery Construction}

Moving from Protocol V1 to Protocol V2 enlarges the gallery by adding
samples from non-target identities across all available views and sequence
types, while leaving the probes and positive templates unchanged. The
resulting performance change therefore reflects the joint effect of a
larger gallery and greater view and sequence-type diversity among its
impostor samples.

The effect of this change depends strongly on both modality and probe
condition. RGB and IR appearance remain comparatively stable under BG, whereas
depth and the event modality are more
sensitive to the enlarged gallery. Under CL, substantial degradation occurs
across nearly all modalities, indicating that the enlarged gallery
particularly amplifies identity ambiguity for CL probes. The magnitude of the
degradation nevertheless varies across modalities, revealing modality-specific
sensitivity to gallery construction under Protocol V2.

\subsubsection{Comparative Analysis of Gait Modalities}

The leading RGB-silhouette method is stable across probe conditions and
protocols: DeepGaitV2-P3D consistently achieves the best Rank-1 and mAP. By
contrast, the best-performing 2D pose method changes with the probe condition:
SkeletonGait performs best under BG, whereas GPGait++ achieves the strongest
results under CL. GPGait++ also consistently
outperforms GPGait for both 2D and 3D pose. With GPGait++ held fixed, however,
estimated 3D pose remains substantially below 2D pose. This gap may reflect
differences in the pose representations and estimation pipelines rather than
an inherent limitation of 3D body structure.

Turning to modality-level comparisons, we use the best-performing method in
each setting when a modality is evaluated with multiple methods. Across both
protocols, RGB and IR appearance lead under BG, whereas LiDAR projected depth
and IR appearance lead under CL. Depth also consistently outperforms RGB
appearance under CL. By contrast, the event modality and both radar modalities
remain comparatively weak under Protocol V2, especially for CL probes.

Comparisons within individual sensing streams further show how
modality-specific processing affects the retained identity information. In the RGB stream,
appearance is more discriminative than the silhouette under BG, whereas the
silhouette performs better under CL. In the IR stream, appearance consistently
outperforms the silhouette; this gap may reflect both the removal of
near-infrared appearance cues during binarization and errors introduced by IR
silhouette segmentation. The preferred encoding also differs between the
LiDAR and radar streams. LiDAR projected depth slightly outperforms its native
point cloud, whereas the radar point cloud is stronger than its projection.
This opposite pattern may reflect greater information loss when sparse radar
returns are projected onto a regular image plane.

Overall, no evaluated modality--model pair is uniformly superior across probe
conditions and gallery protocols. Single-modal discriminability depends on the
probe condition, input representation, and recognition method. Single-modal
evaluation alone, however, cannot determine whether identity descriptors are
compatible across modalities or whether jointly available modalities provide
complementary information. We examine these questions next through cross-modal
and multi-modal recognition.

\subsection{Cross-Modal Recognition}

\subsubsection{Cross-Modal Baseline}

Cross-modal gait recognition matches a probe in one modality against a gallery
represented in another. We evaluate it as directed retrieval, with each
modality serving as the probe and gallery in turn. We adopt
CL-Gait~\cite{guo2024camera} as the task-specific baseline, training one model
for each selected unordered modality pair. Early layers are
modality-specific and deeper layers are shared. The two branches are
jointly optimized using identity classification and a symmetric cross-modal
triplet objective. Symmetry is achieved by constructing triplets in both
directions: each modality alternately supplies anchor samples, while positive
and negative samples are drawn from the other modality. This
objective encourages identity discrimination and cross-modal alignment
simultaneously.

To obtain a controlled comparison across sensing streams, we select four
modalities with a compatible spatial organization: RGB silhouette, IR
silhouette, depth, and LiDAR projected depth. The two silhouettes provide a
shared contour abstraction across visible and infrared sensing, while depth
and LiDAR projected depth provide spatially compatible geometric measurements.
This selection uses one representative from each of four sensing streams
rather than exhaustively pairing all twelve modalities.
The resulting six unordered pairings cover cross-spectrum contour,
contour-to-geometry, and geometry-to-geometry retrieval, corresponding to
twelve directed tasks. In particular, radar projected depth is excluded
because its limited standalone discriminability would make it difficult to
distinguish alignment quality from modality-specific recognition
failure. Radar projected depth is nevertheless included in the subsequent
multi-modal evaluation to test whether it can provide complementary cues when
fused with stronger modalities. All results in this subsection are
evaluated under Protocol V2.

\subsubsection{Pairwise Retrieval}

\begin{table}[t]
	\centering
	\caption{Cross-modal recognition under Protocol V2 (\%). The arrow denotes
		probe-to-gallery retrieval; Bidirectional Avg. is the arithmetic mean of the two
		directions, and Cross-Modal Avg. equally weights the six modality
		pairs. LiDAR denotes LiDAR projected depth.}
	\label{tab:eval_cross_modal}
	\small
	\setlength{\tabcolsep}{8pt}
	\renewcommand{\arraystretch}{1.0}
	\resizebox{0.80\columnwidth}{!}{%
	\begin{tabular}{@{}lcccc@{}}
		\toprule
		\multirow{2}{*}{\textbf{\tabincell{c}{Cross-Modal\\Settings}}} & \multicolumn{2}{c}{\textbf{BG}} &
		\multicolumn{2}{c}{\textbf{CL}} \\
		\cmidrule(lr){2-3}\cmidrule(lr){4-5}
		& R1 & mAP & R1 & mAP \\
		\midrule
		RGB Sil. $\rightarrow$ IR Sil. & 86.0 & 85.3 & 31.0 & 33.3 \\
		IR Sil. $\rightarrow$ RGB Sil. & 83.7 & 83.8 & 28.1 & 31.0 \\
		\textit{Bidirectional Avg.} & 84.9 & 84.5 & 29.5 & 32.1 \\
		\midrule
		RGB Sil. $\rightarrow$ Depth & 51.0 & 51.6 & 18.2 & 20.5 \\
		Depth $\rightarrow$ RGB Sil. & 38.2 & 40.6 & 16.1 & 18.5 \\
		\textit{Bidirectional Avg.} & 44.6 & 46.1 & 17.2 & 19.5 \\
		\midrule
		RGB Sil. $\rightarrow$ LiDAR & 59.6 & 60.4 & 24.8 & 27.4 \\
		LiDAR $\rightarrow$ RGB Sil. & 50.8 & 52.3 & 22.5 & 25.0 \\
		\textit{Bidirectional Avg.} & 55.2 & 56.4 & 23.6 & 26.2 \\
		\midrule
		IR Sil. $\rightarrow$ Depth & 39.8 & 41.2 & 10.6 & 12.9 \\
		Depth $\rightarrow$ IR Sil. & 33.2 & 34.8 & 11.6 & 13.7 \\
		\textit{Bidirectional Avg.} & 36.5 & 38.0 & 11.1 & 13.3 \\
		\midrule
		IR Sil. $\rightarrow$ LiDAR & 52.3 & 53.7 & 17.0 & 19.8 \\
		LiDAR $\rightarrow$ IR Sil. & 47.3 & 47.9 & 17.7 & 19.7 \\
		\textit{Bidirectional Avg.} & 49.8 & 50.8 & 17.3 & 19.7 \\
		\midrule
		Depth $\rightarrow$ LiDAR & 35.1 & 37.8 & 17.5 & 20.3 \\
		LiDAR $\rightarrow$ Depth & 37.3 & 38.8 & 17.8 & 20.3 \\
		\textit{Bidirectional Avg.} & 36.2 & 38.3 & 17.6 & 20.3 \\
		\midrule
		\textbf{\textit{Cross-Modal Avg.}} & 51.2 & 52.3 & 19.4 & 21.9 \\
		\bottomrule
	\end{tabular}}
\vspace{-2ex}
\end{table}

Table~\ref{tab:eval_cross_modal} reports the twelve directed retrieval tasks
together with pair-level and overall averages.
The equal-weight Cross-Modal Avg. reaches $51.2\%$ Rank-1 and $52.3\%$ mAP
under BG. Under CL, it decreases to $19.4\%$ Rank-1 and $21.9\%$ mAP. The
decline across all six modality pairs shows that clothing change further
increases the difficulty of cross-modal matching.

Cross-modal retrieval between RGB and IR silhouettes achieves the strongest
performance, suggesting that their shared contour abstraction facilitates
cross-modal matching. Pairing either silhouette with a geometric modality is
markedly harder. Notably, depth
and LiDAR projected depth achieve
comparatively strong single-modal recognition, but their bidirectional
cross-modal Rank-1 average is only $36.2\%$ under BG and $17.6\%$ under CL.
Shared geometric information and
standalone discriminability are therefore insufficient to guarantee
cross-modal descriptor compatibility. These results demonstrate the difficulty
of cross-modal alignment for the evaluated modality pairs. We next
examine whether modalities provide complementary
identity information through fusion.

\subsection{Multi-Modal Recognition}

\subsubsection{Multi-Modal Baseline}

We evaluate jointly available modalities using the two-modality fusion
strategy of MultiGait++~\cite{jin2025exploring}, implemented with two feature
streams. Each modality is first
processed by a separate shallow encoder, after which the two feature streams
are aggregated for joint identity learning. We use RGB silhouettes as the
anchor because they are widely adopted and provide an appearance-reduced
structural reference based primarily on body shape and motion. Keeping the
anchor fixed provides a consistent reference for comparing the anchor-relative
performance gains produced by different auxiliary modalities, with each
configuration pairing it with one auxiliary input.
The 2D pose coordinates are rendered as joint and limb heatmaps, while LiDAR
and radar denote their projected-depth modalities.

We group the configurations according to the source of the auxiliary
modality. Intra-sensor fusion pairs RGB silhouettes with one of three
RGB-derived auxiliary inputs: 2D pose heatmaps, event sequences, or RGB
appearance. This setup tests whether alternative descriptions of the same
visual signal provide additional information. Inter-sensor fusion pairs the
same anchor with depth, LiDAR projected depth, radar projected depth, IR
appearance, or IR silhouettes to evaluate complementarity across physical
sensing streams. Each configuration is handled by a separately trained expert
model. These pairwise expert settings provide task-specific references for
comparing intra- and inter-sensor complementarity.

\subsubsection{Intra- and Inter-Sensor Fusion}

Table~\ref{tab:eval_multimodal} reports the eight two-modality fusion
configurations, labeled Two-Stream, under Protocol V2 and groups them into
intra- and inter-sensor settings.
The equal-weight Two-Stream Avg. reaches $96.1\%$ Rank-1 and $95.8\%$ mAP
under BG. Under CL, it decreases to $58.2\%$ Rank-1 and $60.6\%$ mAP.

\begin{table}[t]
	\centering
	\caption{Multi-modal recognition under Protocol V2 (\%). Each block first
		lists the RGB-silhouette anchor. Superscripts denote changes relative to
		this anchor; green and red indicate improvements and degradations.
		Two-Stream Avg. equally weights the eight fusion configurations and
		excludes the repeated anchor rows. LiDAR and Radar denote their
		projected-depth modalities.}
	\label{tab:eval_multimodal}
	\setlength{\tabcolsep}{4pt}
	\renewcommand{\arraystretch}{1.0}
	\resizebox{0.88\columnwidth}{!}{%
	\begin{tabular}{@{}lcccc@{}}
		\toprule
		\multirow{2}{*}{\textbf{\tabincell{c}{Multi-Modal}}} &
		\multicolumn{2}{c}{\textbf{BG}} & \multicolumn{2}{c}{\textbf{CL}} \\
		\cmidrule(lr){2-3}\cmidrule(lr){4-5}
		& R1 & mAP & R1 & mAP \\
		\midrule
		\multicolumn{5}{c}{\textbf{Intra-Sensor}} \\
		RGB Sil. & 94.0 & 93.7 & 47.0 & 49.5 \\
		+ Pose
		& 94.6\textsuperscript{\textcolor{green!60!black}{+0.6}}
		& 94.1\textsuperscript{\textcolor{green!60!black}{+0.4}}
		& 52.3\textsuperscript{\textcolor{green!60!black}{+5.3}}
		& 54.5\textsuperscript{\textcolor{green!60!black}{+5.0}} \\
		+ Event
		& 93.8\textsuperscript{\textcolor{red!70!black}{$-$0.2}}
		& 93.4\textsuperscript{\textcolor{red!70!black}{$-$0.3}}
		& 50.5\textsuperscript{\textcolor{green!60!black}{+3.5}}
		& 52.8\textsuperscript{\textcolor{green!60!black}{+3.3}} \\
		+ RGB
		& 99.3\textsuperscript{\textcolor{green!60!black}{+5.3}}
		& 99.4\textsuperscript{\textcolor{green!60!black}{+5.7}}
		& 50.4\textsuperscript{\textcolor{green!60!black}{+3.4}}
		& 53.7\textsuperscript{\textcolor{green!60!black}{+4.2}} \\
		\midrule
		\multicolumn{5}{c}{\textbf{Inter-Sensor}} \\
		RGB Sil. & 94.0 & 93.7 & 47.0 & 49.5 \\
		+ Depth
		& 96.4\textsuperscript{\textcolor{green!60!black}{+2.4}}
		& 96.1\textsuperscript{\textcolor{green!60!black}{+2.4}}
		& 66.5\textsuperscript{\textcolor{green!60!black}{+19.5}}
		& 68.2\textsuperscript{\textcolor{green!60!black}{+18.7}} \\
		+ LiDAR
		& 96.6\textsuperscript{\textcolor{green!60!black}{+2.6}}
		& 96.4\textsuperscript{\textcolor{green!60!black}{+2.7}}
		& 66.7\textsuperscript{\textcolor{green!60!black}{+19.7}}
		& 69.1\textsuperscript{\textcolor{green!60!black}{+19.6}} \\
		+ Radar
		& 94.9\textsuperscript{\textcolor{green!60!black}{+0.9}}
		& 94.6\textsuperscript{\textcolor{green!60!black}{+0.9}}
		& 60.1\textsuperscript{\textcolor{green!60!black}{+13.1}}
		& 62.2\textsuperscript{\textcolor{green!60!black}{+12.7}} \\
		+ IR
		& 98.8\textsuperscript{\textcolor{green!60!black}{+4.8}}
		& 98.8\textsuperscript{\textcolor{green!60!black}{+5.1}}
		& 70.9\textsuperscript{\textcolor{green!60!black}{+23.9}}
		& 73.4\textsuperscript{\textcolor{green!60!black}{+23.9}} \\
		+ IR Sil.
		& 94.4\textsuperscript{\textcolor{green!60!black}{+0.4}}
		& 93.8\textsuperscript{\textcolor{green!60!black}{+0.1}}
		& 48.2\textsuperscript{\textcolor{green!60!black}{+1.2}}
		& 50.6\textsuperscript{\textcolor{green!60!black}{+1.1}} \\
		\midrule
		\textbf{\textit{Two-Stream Avg.}} & 96.1 & 95.8 & 58.2 & 60.6 \\
		\bottomrule
	\end{tabular}}
\vspace{-2ex}
\end{table}

The intra-sensor results reveal condition-dependent complementarity among
alternative representations derived from the RGB stream. The pose and event
modalities contribute substantially more under CL than under BG, with event
fusion slightly degrading the anchor in the latter condition. RGB appearance
exhibits a different pattern: its fusion with the silhouette substantially
improves upon the anchor under BG but only marginally surpasses the already
strong RGB-appearance single-modal baseline; under CL, it clearly outperforms
both the anchor and the RGB-appearance baseline.

All five inter-sensor fusion configurations outperform both the
RGB-silhouette anchor and their corresponding auxiliary single-modal baselines.
Depth, LiDAR projected depth, and IR appearance produce particularly large
anchor-relative gains, indicating substantial complementarity with RGB
contours under the selected expert baseline. Among them, IR appearance gives
the best inter-sensor fusion performance. In contrast, IR silhouettes provide
only limited gains, which may reflect both contour redundancy and the quality
of IR silhouette extraction. Radar projected depth remains useful as an
auxiliary modality despite its weak standalone discriminability, showing that
standalone strength does not fully determine fusion value.

\subsection{Discussion and Key Findings}

\textbf{Discriminability across probe conditions.}
The single-modal results show that no modality is uniformly superior
across the probe conditions. RGB and IR appearance lead under BG,
whereas IR appearance and LiDAR geometry are strongest under CL. Comparisons
among alternative modalities derived from the same sensing stream suggest that
input construction and encoder choice jointly affect recognition.
A unified model should therefore preserve modality-specific cues rather
than impose identical processing on all inputs.

\textbf{Cross-modal compatibility and fusion complementarity.}
Standalone discriminability, cross-modal compatibility, and fusion utility do
not necessarily coincide. Depth and LiDAR projected depth perform
comparatively well in single-modal recognition, yet cross-modal retrieval
between them remains challenging. Conversely, radar projected depth is weak in
standalone recognition but useful as an auxiliary modality under CL,
whereas IR silhouettes provide limited gains and event fusion can introduce
negative transfer under BG. These contrasts motivate fusion mechanisms that
adapt to the available modality combination rather than treating every
auxiliary modality as uniformly beneficial.

\textbf{Scalability of task-specific experts.}
Under the task-specific expert paradigm evaluated above, each individual
modality, cross-modal pair, and fusion configuration is
handled by a separately trained model. As the number of supported
modalities increases, pair-specific models grow quadratically and
subset-specific fusion experts grow combinatorially, making exhaustive model
development, storage, and deployment increasingly impractical. These findings
motivate a unified model that preserves modality-specific cues, aligns
heterogeneous modalities, and adaptively aggregates a variable subset of
the supported modalities within a shared identity space.

\section{Omni-Modal Gait Recognition}
\label{sec:omni_approach}

In this section, we formulate
\emph{Omni-Modal Gait Recognition} to replace a growing collection of
task-specific experts with one jointly trained model. The model operates with
different modality inputs and supports single-modal recognition, cross-modal
recognition, and multi-modal recognition. We instantiate this formulation with
OmniGait++, which combines modality-aware identity encoding with anchor-guided
variable-cardinality temporal fusion.

\subsection{Problem Formulation}

MMGait provides twelve heterogeneous gait modalities whose native data
structures include image sequences, point clouds, and joint coordinates. As a
first step toward fully omni-modal gait recognition, we focus on the nine
modalities that can be expressed as image-like sequences and processed
through a common network interface. Together, these modalities retain
observations from all five sensing streams and span appearance, body contours,
geometry, motion, and body structure. Keeping the input structure and backbone
family comparable allows us to isolate the central challenges of omni-modal
learning---preserving
modality-specific cues, aligning heterogeneous identity descriptors, and
fusing variable-cardinality inputs---without conflating them with architectural
differences between image-, point-, and coordinate-based encoders. We denote
this model-level modality set by $\mathcal{M}$:
\begin{equation}
	\begin{aligned}
		\mathcal{M}=\{&
		\texttt{rgb\_sils}, \texttt{rgb}, \texttt{ir\_sils},
		\texttt{ir},\\
		&\texttt{depth}, \texttt{lidar\_depth},
		\texttt{radar\_depth},\\
		&\texttt{event}, \texttt{pose}\}.
	\end{aligned}
	\label{eq:omni_modality_set}
\end{equation}
Here, \texttt{pose} denotes 2D pose rendered as joint-and-limb heatmaps. Native
LiDAR and radar point clouds are left to extensions with dedicated point-cloud
encoders. Likewise, 3D pose is left to a coordinate-based encoder.

For a modality $m\in\mathcal{M}$, let
$\mathbf{X}^{m}=\{\mathbf{x}^{m}_{t}\}_{t=1}^{T_m}$ denote a gait sequence
of modality-dependent length $T_m$. The
sensing streams in MMGait have different acquisition rates and independently
determined temporal boundaries. We therefore associate modality-specific
sequences by identity, sequence type, and view, but do not require one-to-one
frame correspondence across modalities.

Let $\mathbf{z}_m$ denote the identity descriptor obtained from one modality and
$\mathbf{z}_S$ the fused identity descriptor obtained from a modality subset
$S\subseteq\mathcal{M}$. The same jointly learned model is used in three
operating modes:
\begin{equation}
	\begin{aligned}
		\text{single-modal:}\quad&
		d(\mathbf{z}^{p}_{m},\mathbf{z}^{g}_{m}),\\
		\text{cross-modal:}\quad&
		d(\mathbf{z}^{p}_{m_p},\mathbf{z}^{g}_{m_g}),
		\quad m_p\neq m_g,\\
		\text{multi-modal:}\quad&
		d(\mathbf{z}^{p}_{S},\mathbf{z}^{g}_{S}),
		\quad |S|\geq 2,
	\end{aligned}
	\label{eq:omni_operating_modes}
\end{equation}
where superscripts $p$ and $g$ denote probe and gallery, respectively, and
$d(\cdot,\cdot)$ is the descriptor distance. Cross-modal recognition is
evaluated through directed retrieval because the probe and gallery modalities
play different roles,
whereas multi-modal recognition compares probe and gallery descriptors
constructed from the same modality subset $S$.

This formulation imposes three requirements on a unified model. First, it
should preserve modality-specific physical cues rather than forcing
heterogeneous inputs into an identical low-level representation. Second, it
should make identity descriptors sufficiently compatible for cross-modal
recognition. Third, it should adapt the contribution of each
modality to the composition of $S$, avoiding the assumption that all modalities
are uniformly beneficial. OmniGait++ addresses these requirements through a
modality-specific front-end, a modality-aware shared identity encoder, and an
anchor-guided temporal fusion module.

\begin{figure*}[!ht]
	\centering
	\includegraphics[width=0.96\linewidth]{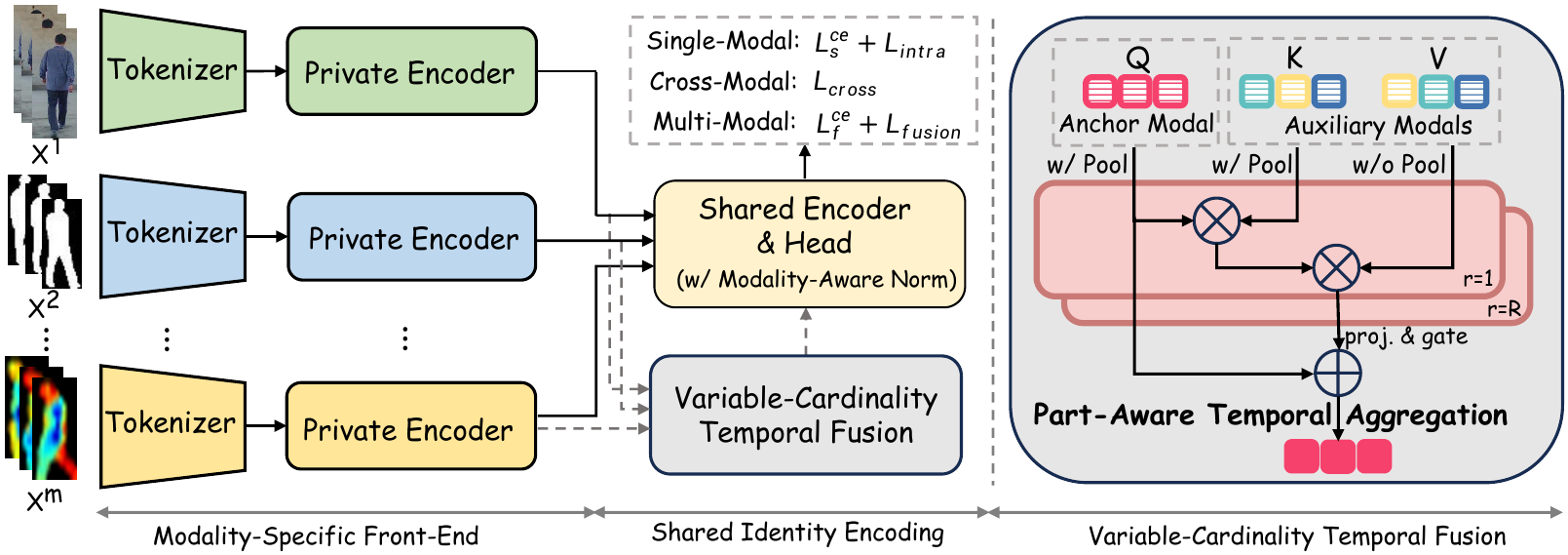}
	\caption{Architecture of OmniGait++. Each modality input $\mathbf{X}^{i}$ is processed
		by a modality-specific tokenizer and private encoder to preserve heterogeneous
		low-level cues. For single- and cross-modal recognition, the private features
		are passed directly to the shared identity encoder and head; for multi-modal
		recognition, anchor-guided temporal fusion first integrates auxiliary feature
		sequences whose number and temporal lengths may vary. Within each horizontal
		region $r\in\{1,\ldots,R\}$, pooled anchor and auxiliary tokens form $Q$ and
		$K$, respectively, while the unpooled auxiliary regional features form $V$;
		the attended context is projected, gated, and added to the anchor
		representation. The loss block summarizes the single-modal and fused
		classification losses $\mathcal{L}^{\mathrm{ce}}_{\mathrm{s}}$ and
		$\mathcal{L}^{\mathrm{ce}}_{\mathrm{f}}$, together with the intra-modal,
		cross-modal, and fusion triplet objectives $\mathcal{L}_{\mathrm{intra}}$,
		$\mathcal{L}_{\mathrm{cross}}$, and $\mathcal{L}_{\mathrm{fusion}}$.}
	\label{fig:omnigait_architecture}
	\vspace{-2ex}
\end{figure*}

\subsection{Overview of OmniGait++}

Figure~\ref{fig:omnigait_architecture} provides an overview of OmniGait++,
which adopts a private-to-shared architecture to separate low-level modality
adaptation from high-level identity learning. For each modality
$m\in\mathcal{M}$, a modality-specific tokenizer $\mathcal{T}_m$, implemented
as an input stem, first adapts the modality's input channel structure to a
common feature format. The backbone then processes the tokenized sequence
through four sequential feature-extraction stages~\cite{fan2023exploring}. We use
$s\in\{1,2,3,4\}$ to denote the last private stage: Stages~1--$s$
form the modality-specific encoder $E^{\mathrm{pri}}_{m,1:s}$, while the
remaining stages, if any, form the shared encoder $E^{\mathrm{sh}}_s$. Unless
otherwise stated, we use $s=3$. When $s=4$, no shared backbone stage remains,
and $E^{\mathrm{sh}}_4$ is the identity mapping. The private encoder produces
$\mathbf{U}^{m}\in\mathbb{R}^{C\times T_m\times H\times W}$ at this boundary
while retaining the native temporal length $T_m$.

For a single-modal input, the descriptor pathway is
\begin{equation}
	\mathbf{U}^{m}
		=E^{\mathrm{pri}}_{m,1:s}
		\bigl(\mathcal{T}_m(\mathbf{X}^{m})\bigr),
	\qquad
	\mathbf{z}_m
		=\mathcal{H}\bigl(E^{\mathrm{sh}}_s
		(\mathbf{U}^{m};m)\bigr),
	\label{eq:omnigait_single_path}
\end{equation}
where $\mathcal{H}$ comprises temporal pooling, horizontal part pooling, and
part-wise feature projection following common
gait-recognition practice~\cite{fan2023opengait}. The argument after the
semicolon selects the modality-aware normalization branch, with $m$ denoting
the branch for the current single modality. Thus, the private prefix preserves
modality-dependent cues, while the shared encoder and descriptor head map them
through a common high-level parameterization.

When multiple modalities are jointly available, let $S\subseteq\mathcal{M}$
be an evaluated fusion subset containing the RGB-silhouette anchor
$a=\texttt{rgb\_sils}$. OmniGait++ inserts a single anchor-guided fusion module
$\Phi$ at the private-to-shared boundary:
\begin{equation}
	\begin{aligned}
		\widetilde{\mathbf{U}}^{S}
			&=\Phi\bigl(
				\mathbf{U}^{a},
				\{\mathbf{U}^{m}\}_{m\in S\setminus\{a\}}
			\bigr),\\
		\mathbf{z}_{S}
			&=\mathcal{H}\bigl(
				E^{\mathrm{sh}}_s(\widetilde{\mathbf{U}}^{S};
				\mathrm{fusion})
			\bigr).
	\end{aligned}
	\label{eq:omnigait_fusion_path}
\end{equation}
The module aggregates the auxiliary private features into the anchor
representation, producing $\widetilde{\mathbf{U}}^{S}$ with the anchor timeline
and spatial layout. The fused feature then passes through the same shared
identity encoder and descriptor head as the single-modal feature, while
$\mathrm{fusion}$ selects a dedicated normalization branch. Sharing both
$\Phi$ and the downstream pathway allows one model to process subsets with
different compositions and cardinalities without configuration-specific
networks.

These two descriptor pathways instantiate the three operating modes in
Eq.~\eqref{eq:omni_operating_modes}. Single-modal recognition compares
descriptors $\mathbf{z}_m$ from the same modality. Cross-modal recognition
independently encodes the probe and gallery modalities and directly compares
$\mathbf{z}_{m_p}$ with $\mathbf{z}_{m_g}$ without invoking $\Phi$.
Multi-modal recognition applies $\Phi$ to the same subset $S$ on the probe and
gallery sides and compares their fused descriptors $\mathbf{z}_S$. All
branches are learned jointly with intra-modal, cross-modal, and fusion-specific
supervision.

\textbf{From OmniGait to OmniGait++.}
The primary extension from the preliminary OmniGait
model~\cite{wan2026mmgait} to OmniGait++ is fusion scalability. OmniGait
restricts each fusion input to an RGB-silhouette anchor and exactly one
auxiliary modality, and therefore supports only fixed-cardinality pairwise
fusion. In contrast, OmniGait++ represents the auxiliary modalities as a
variable-size set and uses one shared temporal attention module to aggregate
anchor-containing subsets with different compositions and cardinalities. The
same fusion pathway can therefore accommodate different numbers and types of
available modalities without changing the fusion architecture. In addition,
OmniGait uses a shallow modality-specific front-end followed by shared backbone
stages, whereas OmniGait++ extends modality-specific processing to a
configurable private-to-shared boundary and applies modality-aware
normalization in the shared encoder.

\subsection{Modality-Aware Shared Identity Encoder}

Heterogeneous gait observations exhibit substantially different low-level
statistics, yet their identity descriptors must remain comparable across
modalities. OmniGait++ addresses this tension by decomposing its encoder into a
modality-specific front-end and a shared identity-encoding pathway. The former
absorbs input-dependent appearance and geometry, while the latter promotes a
common high-level representation through shared convolutional and projection
parameters. Modality-aware normalization is inserted into the shared pathway
to prevent incompatible modality statistics from being indiscriminately mixed.

\subsubsection{Modality-Specific Front-End}

The modalities in $\mathcal{M}$ differ in channel structure and physical
meaning. Silhouettes contain binary contours; RGB/IR inputs retain appearance;
projected depth encodes geometry; the event modality emphasizes motion; pose
heatmaps encode body structure. Applying a common stem directly to these inputs
can force the earliest layers to account simultaneously for incompatible signal
distributions. We instead associate each modality $m$ with an independent
tokenizer $\mathcal{T}_m$, implemented as an input stem that maps its frames to
a common feature format.

The tokenized sequence is subsequently processed through Stage~$s$ by a
modality-specific private encoder, producing
the private feature $\mathbf{U}^{m}$ in
Eq.~\eqref{eq:omnigait_single_path}. All convolutional and normalization
parameters in this private prefix are independently learned for each modality,
preserving modality-dependent patterns before the features enter either the
shared encoder or the fusion module. Meanwhile, the common output format
ensures that all $\mathbf{U}^{m}$ have compatible channel and spatial
dimensions while retaining their native sequence lengths $T_m$.

\subsubsection{Shared Identity Encoding}

After private Stage~$s$, both single-modal features $\mathbf{U}^{m}$ and fused
features $\widetilde{\mathbf{U}}^{S}$ are processed by the shared encoder
$E^{\mathrm{sh}}_s$. Its convolutional parameters are shared across
modalities and fusion subsets, encouraging identity-related structures to be
represented through a common set of high-level transformations.

To obtain sequence-level descriptors, we first apply temporal max pooling over
the valid frames and then divide the resulting feature map into horizontal
parts~\cite{fan2023opengait}. Within each part, the average- and max-pooled responses are summed to
retain both distributed and salient information. Part-specific linear projections,
shared across all modalities and fusion subsets, produce the final
$\mathbf{z}_m$ and $\mathbf{z}_S$. Sharing the encoder and part-specific
projections across modalities and fusion subsets gives the three operating
modes a common high-level parameterization.

\subsubsection{Modality-Aware Normalization}

Conventional batch normalization in the shared pathway would estimate one set
of statistics from mixed single-modal and fused features. Such mixed statistics can
obscure modality-dependent distributions. Inspired by the general principle of
domain-specific batch normalization~\cite{chang2019domain}, we therefore use a
modality-aware variant in the shared encoder. For a normalization key
$q\in\mathcal{M}\cup\{\mathrm{fusion}\}$, the operation is written as
\begin{equation}
	\mathcal{N}(\mathbf{x};q)
		=\gamma_q
		\frac{\mathbf{x}-\mu_q}
		{\sqrt{\sigma_q^2+\epsilon}}
		+\beta_q,
	\label{eq:modality_aware_norm}
\end{equation}
where $(\mu_q,\sigma_q^2)$ are running statistics specific to $q$. For the nine
single modalities, the affine parameters are tied,
$(\gamma_q,\beta_q)=(\gamma_{\mathrm{single}},
\beta_{\mathrm{single}})$ for $q\in\mathcal{M}$, while the fusion branch keeps
its own affine parameters. This design preserves modality-dependent feature
calibration without relaxing the shared convolutional representation into
independent encoders. The same mechanism is applied to the spatial and temporal
normalization layers of the shared P3D blocks~\cite{qiu2017learning}. In contrast, the tokenizer and
private stages maintain independently learned normalization parameters for
each modality.

Following the BNNeck design~\cite{luo2019bag}, we further use modality-aware
BNNecks for identity classification. Each
single modality maintains its own BN statistics but uses a common
single-modal classifier, whereas all fused descriptors use a dedicated
fusion BN branch and classifier:
\begin{equation}
	\boldsymbol{\ell}_{m}
		=C_{\mathrm{single}}\bigl(B_m(\mathbf{z}_m)\bigr),~
	\boldsymbol{\ell}_{S}
		=C_{\mathrm{fusion}}\bigl(B_{\mathrm{fusion}}(\mathbf{z}_S)\bigr).
	\label{eq:modality_aware_bnneck}
\end{equation}
The shared single-modal classifier provides common identity supervision across
modalities, while the separate fusion classifier accommodates the
distribution shift introduced by feature aggregation. The pre-BN descriptors
$\mathbf{z}_m$ and $\mathbf{z}_S$ remain the inputs to metric learning and
retrieval.

\subsection{Variable-Cardinality Temporal Fusion}

Conventional channel-wise feature concatenation typically assumes a fixed set
of input modalities and therefore requires the aggregation layer to be
redesigned when the modality composition changes. It is also poorly suited to MMGait,
whose sensing streams have different sequence lengths and no prescribed
frame-level correspondence. We instead formulate fusion as anchor-guided
temporal aggregation over a variable-size auxiliary set. This formulation
fixes the output reference while allowing the number, type, and temporal extent
of auxiliary observations to change across operating settings.

\subsubsection{Anchor-Guided Fusion Formulation}

Let $S$ be an evaluated fusion subset containing
$a=\texttt{rgb\_sils}$ as the anchor. RGB silhouettes are used as the anchor
because they suppress appearance variation while retaining body contours,
providing a consistent spatial reference across fusion configurations. The
anchor-guided temporal fusion module $\Phi$ maps the anchor feature
$\mathbf{U}^{a}$, together with the auxiliary set
$\{\mathbf{U}^{m}\}_{m\in S\setminus\{a\}}$, to the fused representation
$\widetilde{\mathbf{U}}^{S}$. This representation has the same shape as
$\mathbf{U}^{a}$: the anchor determines the output timeline and feature layout,
whereas the remaining modalities provide auxiliary evidence.
Because the auxiliary inputs are represented as a set rather than assigned to
fixed branches, the same $\Phi$ is reused for all anchor-containing subsets.

\subsubsection{Part-Aware Temporal Aggregation}

Gait cues are not distributed uniformly over the human body, and the
reliability of corresponding regions can differ across modalities. We
therefore partition the feature height into $R$ non-overlapping horizontal
regions $\{\Omega_r\}_{r=1}^{R}$ and perform temporal aggregation separately
within each region. These fusion regions are independent of the horizontal
parts used by the output head. For modality $m$, spatial averaging within
region $\Omega_r$ produces a sequence of part-aware tokens:
\begin{equation}
	\mathbf{T}^{m}_{r}
		=\operatorname{Pool}_{\Omega_r}
		(\mathbf{U}^{m})+\mathbf{e}_m,
	\qquad
	\mathbf{T}^{m}_{r}\in\mathbb{R}^{T_m\times C},
	\label{eq:fusion_part_tokens}
\end{equation}
where $\mathbf{e}_m$ is a learnable modality embedding. It identifies the
source of each token after auxiliary modalities are gathered into a common
sequence.

For each region, the anchor tokens form the queries, while tokens from all
auxiliary modalities are concatenated along the joint modality--time dimension to
form the keys. Let $\mathbf{U}^{m}_{r}$ denote the corresponding full feature
map in region $\Omega_r$. The queries, keys, and values are constructed as
\begin{equation}
	\begin{aligned}
		\mathbf{Q}_r
			&=\mathcal{P}^{Q}_{r}
			\bigl(\mathcal{N}_{r}(\mathbf{T}^{a}_{r})\bigr),\\
		\mathbf{K}_r
			&=\underset{m\in S\setminus\{a\}}{\operatorname{Concat}}
			\mathcal{P}^{K}_{r}
			\bigl(\mathcal{N}_{r}(\mathbf{T}^{m}_{r})\bigr),\\
		\mathbf{V}_r
			&=\underset{m\in S\setminus\{a\}}{\operatorname{Concat}}
			\mathcal{P}^{V}_{r}
			\bigl(\mathcal{N}^{V}_{r}(\mathbf{U}^{m}_{r})\bigr),
	\end{aligned}
	\label{eq:fusion_queries_keys}
\end{equation}
where $\mathcal{N}_{r}$ and $\mathcal{P}_{r}$ denote the corresponding
normalization and projection operators, respectively. Crucially, the anchor is
excluded from both the key and value sets and participates in attention only
as the query. The values retain the complete feature map inside each horizontal
region rather than only its pooled token. Consequently, token similarity
determines how full local feature maps from auxiliary modalities and times
contribute to each anchor position.

Aggregation is implemented with multi-head normalized
attention~\cite{vaswani2017attention}. For region
$r$, its attention weights can be written compactly as
\begin{equation}
	\mathbf{A}_r
		=\operatorname{softmax}\left(
			\rho_r\,
			\operatorname{cos}(\mathbf{Q}_r,\mathbf{K}_r)
		\right),
	\label{eq:fusion_attention}
\end{equation}
where $\rho_r$ is a positive learnable scale. Cosine similarity reduces
sensitivity to modality-dependent feature magnitudes, while multiple heads
allow different relationships among modalities to be captured in parallel.
The attended values produce an anchor-conditioned local context
$\mathbf{C}_r=\mathbf{A}_r\mathbf{V}_r$. Because auxiliary observations are
concatenated along the joint modality--time dimension, this aggregation
accommodates unequal sequence lengths without requiring frame-level
correspondence. The number of anchor queries determines the temporal length of
$\mathbf{C}_r$, thereby aligning it with the anchor timeline. Although
conditioned on the anchor queries, $\mathbf{C}_r$ aggregates only auxiliary
values. Contexts from all regions are concatenated in their original vertical
order and passed through a spatial projection $\mathcal{O}$ to form an
auxiliary correction. The original anchor feature bypasses value aggregation
and is retained through the residual path:
\begin{equation}
	\widetilde{\mathbf{U}}^{S}
		=\mathbf{U}^{a}
		+\alpha\,
		\mathcal{O}\left(
			\operatorname{Concat}_{H}
			(\mathbf{C}_1,\ldots,\mathbf{C}_R)
		\right),
	\label{eq:fusion_residual_output}
\end{equation}
where $\alpha\in\mathbb{R}$ is a learnable scalar initialized to $0.1$. This
exclusion prevents the anchor's self-response from competing with auxiliary
evidence in the key--value set. The residual connection preserves the strong
single-modal reference, while $\alpha$ controls the contribution of the
auxiliary correction when an auxiliary modality is noisy or weakly
complementary.

\subsection{Unified Identity Learning and Inference}

Having defined the single-modal and fused descriptors, we optimize their
geometry in the shared identity space with objectives aligned to the three
operating modes. We first learn discriminative features within each modality,
then enforce directed cross-modal alignment, and finally supervise the fused
descriptors. The resulting descriptors are then directly reused by the
corresponding inference modes.

\subsubsection{Single-Modal Identity Learning}

The nine single-modal branches are trained jointly using classification and
metric objectives. As described in
Eq.~\eqref{eq:modality_aware_bnneck}, their modality-aware BNNecks share one
classifier. This exposes the classifier to every modality and encourages
their normalized features to predict identity labels using a common decision
space. We apply label-smoothed cross-entropy to the part-wise logits of every
modality and average their contributions equally, denoting the resulting
objective by $\mathcal{L}^{\mathrm{ce}}_{\mathrm{s}}$.

Classification alone does not explicitly constrain the retrieval geometry of
the pre-BN descriptors. We therefore additionally apply a batch-all triplet
objective~\cite{hermans2017defense} independently within each modality and
average the nine losses to obtain $\mathcal{L}_{\mathrm{intra}}$. This
group-wise computation gives every
modality equal weight and prevents cross-modal distances from being
introduced into the intra-modal objective.

\subsubsection{Cross-Modal Identity Alignment}

The shared classifier provides an implicit connection among modalities, but
does not directly optimize distances between probe and gallery descriptors
from different modalities. We impose explicit cross-modal metric supervision on
the four modalities used by the directed cross-modal retrieval protocol:
\texttt{rgb\_sils}, \texttt{ir\_sils}, \texttt{depth}, and
\texttt{lidar\_depth}. For each ordered pair, descriptors from the first
modality serve as anchors, while same-identity and different-identity
descriptors from the second modality form positives and negatives,
respectively. A batch-all triplet loss is computed independently for every
horizontal part. Reversing a pair changes the modality that supplies the
anchors and therefore produces a distinct optimization direction. We optimize
all twelve directions formed by the four modalities and average their losses
equally to obtain $\mathcal{L}_{\mathrm{cross}}$. This directly matches the
directed cross-modal retrieval protocol and prevents any modality pair from
dominating the shared identity space.

\subsubsection{Multi-Modal Fusion Learning}

The distribution of a fused descriptor differs from that of a single-modal
descriptor. Fused descriptors are therefore supervised through the
dedicated fusion BNNeck and classifier in
Eq.~\eqref{eq:modality_aware_bnneck}. The fusion configurations processed in
each training iteration share the fusion classifier, and their label-smoothed
cross-entropy losses are averaged to form
$\mathcal{L}^{\mathrm{ce}}_{\mathrm{f}}$. Sharing this classifier encourages
fused descriptors from subsets with different compositions and cardinalities
to remain identity-discriminative under a common set of decision boundaries.

To preserve the retrieval geometry of each fusion setting, batch-all triplets
are formed separately within each training fusion configuration and then
averaged across configurations, yielding $\mathcal{L}_{\mathrm{fusion}}$. This
keeps metric learning consistent with multi-modal evaluation, where probe and
gallery inputs are constructed from the same modality subset.

\subsubsection{Overall Objective}

The complete learning objective combines the two classification terms and
three task-specific metric terms:
\begin{equation}
	\mathcal{L}
		=\mathcal{L}^{\mathrm{ce}}_{\mathrm{s}}
		+\mathcal{L}^{\mathrm{ce}}_{\mathrm{f}}
		+\mathcal{L}_{\mathrm{intra}}
		+\mathcal{L}_{\mathrm{cross}}
		+\mathcal{L}_{\mathrm{fusion}}.
	\label{eq:omnigait_total_loss}
\end{equation}
All terms are equally weighted, and a common margin is used for the three
triplet objectives. Computing each term group-wise before averaging ensures
that optimization is balanced over modalities, retrieval directions, and
fusion subsets rather than being biased toward the task containing the largest
number of descriptors.

\subsubsection{Unified Inference}
At inference, the identity classifiers are discarded and retrieval uses
Euclidean distance between pre-BN descriptors. Given one modality $m$,
OmniGait++ produces $\mathbf{z}_m$ for single-modal recognition. For cross-modal
recognition, it independently encodes the probe and gallery modality inputs and
performs directed retrieval by comparing $\mathbf{z}_{m_p}$ with
$\mathbf{z}_{m_g}$ without invoking fusion. Given an
anchor-containing subset $S$, multi-modal recognition produces
$\mathbf{z}_S$ and uses the same modality composition for probe and gallery.
The same checkpoint therefore supports all three modes without task-specific
fine-tuning, score-level fusion, or separately trained networks.

\section{Experiments on Omni-Modal Gait Recognition}
\label{sec:experiments}

\begin{table*}[!t]
	\centering
	\caption{Multi-modal fusion-to-fusion evaluation registry. Every set contains
		the RGB-silhouette anchor and is used for both probe and gallery fusion;
		group labels describe the auxiliary modalities.}
	\label{tab:multimodal_fusion_registry}
	\renewcommand{\arraystretch}{1.0}
	\setlength{\tabcolsep}{4pt}
	\resizebox{0.80\textwidth}{!}{%
		\begin{tabular}{@{}ccc@{}}
			\toprule
			\textbf{Group} &
			\tabincell{c}{\textbf{Fusion Set}} &
			\textbf{Diagnostic Role} \\
			\midrule
			\multirow{8}{*}{Two-Stream} & RGB Sil. + Depth &
			Dense depth geometry complements visible contours. \\
			& RGB Sil. + IR Sil. &
			Cross-spectrum contour consistency. \\
			& RGB Sil. + LiDAR &
			Active 3D structure from LiDAR projected depth. \\
			& RGB Sil. + Radar &
			Radar geometry tests robustness beyond optical sensing. \\
			& RGB Sil. + Event &
			Motion-sensitive evidence from temporal changes. \\
			& RGB Sil. + Pose &
			Structural body layout from pose heatmaps. \\
			& RGB Sil. + RGB &
			Raw RGB appearance beyond silhouette extraction. \\
			& RGB Sil. + IR &
			Spectrum-dependent IR appearance complementarity. \\
			\midrule
			\multirow{3}{*}{Within-Group} & RGB Sil. + RGB + IR Sil. + IR &
			Visible/infrared appearance and contour fusion. \\
			& RGB Sil. + Depth + LiDAR + Radar &
			Range-based geometry modalities fused with anchor. \\
			& RGB Sil. + Event + Pose &
			Motion and pose fusion without range sensors. \\
			\midrule
			\multirow{5}{*}{Across-Group} & RGB Sil. + IR Sil. + Depth &
			Visible contour, IR contour, and metric geometry. \\
			& RGB Sil. + RGB + Depth &
			Contour, raw appearance, and camera geometry. \\
			& RGB Sil. + Event + Depth &
			Contour, motion, and camera geometry. \\
			& RGB Sil. + Pose + LiDAR &
			Contour, pose structure, and LiDAR geometry. \\
			& RGB Sil. + Event + Radar &
			Contour, motion, and radar-derived geometry. \\
			\midrule
			\multirow{5}{*}{Omni-Fusion} & \tabincell{c}{RGB Sil. + Depth + IR Sil. \\ + LiDAR + Radar} &
			Omni-5 contour-and-geometry fusion subset. \\
			& \tabincell{c}{RGB Sil. + RGB + IR Sil. + IR + Depth \\ + LiDAR + Radar} &
			Omni-7, adding appearance modalities to Omni-5. \\
			& \tabincell{c}{RGB Sil. + RGB + IR Sil. + IR + Depth \\ + LiDAR + Radar + Event + Pose} &
			Omni-9 full fusion over all available modalities. \\
			\bottomrule
	\end{tabular}}
	\vspace{-2ex}
\end{table*}

\subsection{Experimental Setup}
\label{sec:experimental_setup}

All main OmniGait++ results in Section~\ref{sec:performance_evaluation} use
\emph{one} jointly trained checkpoint for single-modal recognition,
cross-modal recognition, and multi-modal recognition, without task-specific
fine-tuning or additional checkpoints.

\subsubsection{Evaluation Settings and Metrics}
\label{sec:experimental_protocols}

Unless otherwise stated, all results follow Protocol V2 with the
impostor-augmented gallery defined in Section~\ref{sec:mmgait_protocols}.
The omni-modal experiments reuse the same probe, positive-template, and
impostor definitions as the expert study. Rank-1 accuracy and mAP are reported
separately for BG and CL after uniformly averaging the off-diagonal view
pairs.

\subsubsection{Operating Modes and Evaluation Tasks}
\label{sec:evaluation_tasks}

The three operating modes in
Eq.~\eqref{eq:omni_operating_modes} are instantiated as follows.
\textbf{Single-modal recognition} compares probe and gallery descriptors from
the same modality, yielding nine tasks. \textbf{Cross-modal recognition}
performs retrieval in both directions over the six unordered pairs among RGB
silhouettes, IR silhouettes, depth, and LiDAR projected depth, yielding twelve
directed tasks without invoking the fusion module. \textbf{Multi-modal recognition}
uses a fusion-to-fusion protocol in which probe and gallery are encoded from
the same modality subset $S$.

\subsubsection{Fusion Configuration Registry}
\label{sec:fusion_registry}

To evaluate variable-cardinality fusion systematically, we define the 19
configurations in Table~\ref{tab:multimodal_fusion_registry} as a structured
diagnostic registry. Every configuration contains the RGB-silhouette anchor;
therefore, the group labels refer only to the relationships among the eight
auxiliary modalities. We organize these auxiliaries into three cue families:
appearance/contour (RGB, IR silhouette, and IR), geometry (depth, LiDAR
projected depth, and radar projected depth), and motion/structure (event and
pose).

The registry comprises four diagnostic groups. Two-Stream pairs the anchor
with exactly one auxiliary and exhausts all eight choices, isolating the
contribution of each auxiliary modality. Within-Group combines the anchor with
all auxiliaries from one cue family, yielding three family-level
configurations. Across-Group combines two auxiliaries from different cue
families while fixing the total fusion cardinality at three; each configuration
pairs a geometry modality with a contour, appearance, motion, or structure
modality. These five configurations are representative rather than exhaustive.
Finally, the nested Omni-5, Omni-7, and Omni-9 configurations progressively
extend contour-and-geometry fusion with appearance and then motion/structure
cues. This organization supports complementary analyses of single-auxiliary
utility, within-family aggregation, cross-family complementarity, and scaling
to broader modality subsets.

\begin{table}[t]
	\centering
	\caption{Single-modal results of OmniGait++ under Protocol V2.
		Results are reported as Rank-1 accuracy and mAP (\%).}
	\label{tab:omni_single_modal}
	\small
	\setlength{\tabcolsep}{10pt}
	\renewcommand{\arraystretch}{1.0}
	\resizebox{0.75\columnwidth}{!}{%
		\begin{tabular}{@{}lcccc@{}}
			\toprule
			\multirow{2}{*}{\textbf{Modality}} & \multicolumn{2}{c}{\textbf{BG}} &
			\multicolumn{2}{c}{\textbf{CL}} \\
			\cmidrule(lr){2-3}\cmidrule(lr){4-5}
			& R1 & mAP & R1 & mAP \\
			\midrule
			RGB Sil. & 90.6 & 90.3 & 37.8 & 40.6 \\
			RGB & 99.0 & 99.0 & 45.3 & 48.8 \\
			IR Sil. & 79.8 & 79.2 & 20.8 & 23.5 \\
			IR & 96.8 & 96.9 & 55.7 & 59.3 \\
			Depth & 69.2 & 70.3 & 32.3 & 35.3 \\
			LiDAR & 81.7 & 82.3 & 42.1 & 45.2 \\
			Radar & 1.5 & 2.4 & 0.9 & 1.7 \\
			Event & 21.6 & 24.0 & 3.7 & 5.0 \\
			Pose & 49.0 & 49.7 & 20.3 & 22.5 \\
			\midrule
			\textbf{\textit{Average}} & 65.5 & 66.0 & 28.8 & 31.3 \\
			\bottomrule
	\end{tabular}}
	\vspace{-2ex}
\end{table}

\begin{table}[t]
	\centering
	\caption{Cross-modal recognition of OmniGait++ under Protocol V2 (\%). The arrow denotes
		probe-to-gallery retrieval; Bidirectional Avg. is the arithmetic mean of the two
		directions, and Cross-Modal Avg. equally weights the six modality pairs.
		LiDAR denotes LiDAR projected depth.}
	\label{tab:omni_cross_modal}
	\small
	\setlength{\tabcolsep}{8pt}
	\renewcommand{\arraystretch}{1.0}
	\resizebox{0.80\columnwidth}{!}{%
		\begin{tabular}{@{}lcccc@{}}
			\toprule
			\multirow{2}{*}{\textbf{\tabincell{c}{Cross-Modal\\Settings}}} & \multicolumn{2}{c}{\textbf{BG}} &
			\multicolumn{2}{c}{\textbf{CL}} \\
			\cmidrule(lr){2-3}\cmidrule(lr){4-5}
			& R1 & mAP & R1 & mAP \\
			\midrule
			RGB Sil. $\rightarrow$ IR Sil. & 79.1 & 78.2 & 23.9 & 26.3 \\
			IR Sil. $\rightarrow$ RGB Sil. & 77.0 & 77.5 & 22.5 & 25.5 \\
			\textit{Bidirectional Avg.} & 78.0 & 77.8 & 23.2 & 25.9 \\
			\midrule
			RGB Sil. $\rightarrow$ Depth & 59.6 & 60.3 & 19.1 & 21.8 \\
			Depth $\rightarrow$ RGB Sil. & 46.2 & 48.3 & 17.3 & 19.9 \\
			\textit{Bidirectional Avg.} & 52.9 & 54.3 & 18.2 & 20.8 \\
			\midrule
			RGB Sil. $\rightarrow$ LiDAR & 66.1 & 66.6 & 25.3 & 27.9 \\
			LiDAR $\rightarrow$ RGB Sil. & 55.7 & 56.6 & 22.0 & 24.7 \\
			\textit{Bidirectional Avg.} & 60.9 & 61.6 & 23.7 & 26.3 \\
			\midrule
			IR Sil. $\rightarrow$ Depth & 50.5 & 51.8 & 13.6 & 16.3 \\
			Depth $\rightarrow$ IR Sil. & 39.4 & 41.1 & 12.9 & 15.1 \\
			\textit{Bidirectional Avg.} & 45.0 & 46.5 & 13.3 & 15.7 \\
			\midrule
			IR Sil. $\rightarrow$ LiDAR & 59.7 & 60.6 & 18.7 & 21.6 \\
			LiDAR $\rightarrow$ IR Sil. & 50.7 & 51.2 & 17.3 & 19.5 \\
			\textit{Bidirectional Avg.} & 55.2 & 55.9 & 18.0 & 20.6 \\
			\midrule
			Depth $\rightarrow$ LiDAR & 50.7 & 52.1 & 23.2 & 25.9 \\
			LiDAR $\rightarrow$ Depth & 49.0 & 50.2 & 20.6 & 23.5 \\
			\textit{Bidirectional Avg.} & 49.9 & 51.2 & 21.9 & 24.7 \\
			\midrule
			\textbf{\textit{Cross-Modal Avg.}} & 57.0 & 57.9 & 19.7 & 22.3 \\
			\bottomrule
	\end{tabular}}
	\vspace{-2ex}
\end{table}

\begin{table}[t]
	\centering
	\caption{Variable-cardinality multi-modal fusion results of OmniGait++
		under Protocol V2. Results are Rank-1 accuracy and mAP (\%), with each
		group followed by its average. Omni-5 fuses RGB silhouettes with IR
		silhouettes, depth, LiDAR, and radar; Omni-7 adds RGB and IR appearance;
		Omni-9 uses all nine image-like modalities.}
	\label{tab:omni_multi_modal}
	\footnotesize
	\renewcommand{\arraystretch}{1.0}
	\setlength{\tabcolsep}{4pt}
	\resizebox{0.88\columnwidth}{!}{%
		\begin{tabular}{@{}>{\raggedright\arraybackslash}p{3.9cm} cccc@{}}
			\toprule
			\tabincell{c}{\textbf{Fusion Set}} &
			\multicolumn{2}{c}{\textbf{BG}} & \multicolumn{2}{c}{\textbf{CL}} \\
			\cmidrule(lr){2-3}\cmidrule(lr){4-5}
			& R1 & mAP & R1 & mAP \\
			\midrule
			RGB Sil.+Depth & 95.7 & 95.4 & 57.5 & 59.9 \\
			RGB Sil.+IR Sil. & 93.4 & 93.1 & 42.7 & 45.6 \\
			RGB Sil.+LiDAR & 96.7 & 96.7 & 61.8 & 64.5 \\
			RGB Sil.+Radar & 93.5 & 93.2 & 54.7 & 57.3 \\
			RGB Sil.+Event & 93.2 & 93.0 & 45.4 & 48.2 \\
			RGB Sil.+Pose & 92.0 & 91.5 & 46.4 & 48.9 \\
			RGB Sil.+RGB & 99.7 & 99.7 & 64.6 & 67.5 \\
			RGB Sil.+IR & 99.4 & 99.4 & 75.4 & 77.6 \\
			\midrule
			\textbf{\textit{Two-Stream Avg.}} & 95.5 & 95.3 & 56.1 & 58.7 \\
			\midrule
			RGB Sil.+RGB+IR Sil.+IR & 99.8 & 99.8 & 76.1 & 78.2 \\
			\tabincell{l}{RGB Sil.+Depth+LiDAR+Radar} & 98.6 & 98.5 & 75.1 & 77.1 \\
			RGB Sil.+Event+Pose & 94.3 & 93.9 & 50.8 & 53.4 \\
			\midrule
			\textbf{\textit{Within-Group Avg.}} & 97.6 & 97.4 & 67.3 & 69.5 \\
			\midrule
			RGB Sil.+IR Sil.+Depth & 97.0 & 96.7 & 59.5 & 61.7 \\
			RGB Sil.+RGB+Depth & 99.7 & 99.7 & 70.5 & 72.8 \\
			RGB Sil.+Event+Depth & 96.4 & 96.1 & 58.3 & 60.6 \\
			RGB Sil.+Pose+LiDAR & 97.2 & 97.0 & 65.2 & 67.5 \\
			RGB Sil.+Event+Radar & 95.5 & 95.3 & 57.4 & 60.0 \\
			\midrule
			\textbf{\textit{Across-Group Avg.}} & 97.1 & 97.0 & 62.2 & 64.5 \\
			\midrule
			Omni-5 & 98.8 & 98.7 & 75.9 & 77.7 \\
			Omni-7 & 99.8 & 99.9 & 83.0 & 84.7 \\
			Omni-9 & 99.8 & 99.8 & 83.5 & 85.2 \\
			\midrule
			\textbf{\textit{Omni-Fusion Avg.}} & 99.5 & 99.5 & 80.8 & 82.5 \\
			\bottomrule
	\end{tabular}}
	\vspace{-2ex}
\end{table}

\subsubsection{Implementation Details}
\label{sec:implementation_details}

OmniGait++ builds on the DeepGaitV2-P3D architecture~\cite{fan2023exploring},
using a modality-specific tokenizer followed by a private-to-shared
backbone.\footnote{Code, pretrained models, and instructions for requesting
access to MMGait will be available at
\url{https://github.com/BNU-IVC/MMGait}.}
By default, Stages~1--3 form the modality-specific private encoder,
after which a shared Stage-4 identity encoder, temporal max pooling, horizontal
part pooling, and part-wise projections produce the retrieval descriptors.
Fusion is performed after private Stage~3 and before shared Stage~4. The fusion
module uses $R{=}8$ horizontal fusion regions, eight attention heads, cosine similarity,
learnable modality embeddings, and residual correction. Modality-aware normalization is applied in
both the shared encoder and the BNNecks.

Each mini-batch contains 8
identities and 4 sequences per identity. We sample 30 ordered frames
and apply random perspective, horizontal flipping, and rotation, each with
probability $0.2$. Optimization uses SGD with learning rate $0.1$, momentum
$0.9$, and weight decay $5{\times}10^{-4}$, decayed at 20k, 40k, and 50k
iterations for a total of 60k iterations. The five loss terms in
Eq.~\eqref{eq:omnigait_total_loss} are equally weighted; the triplet margin is
$0.2$. Cross-modal triplets are computed over the twelve directed tasks
defined above. In each iteration, eight fusion subsets are
sampled by drawing two configurations from each of the four registry groups,
while evaluation always uses the full 19-set registry. At inference, the
classifiers are discarded and Euclidean distance is computed on the pre-BN
descriptors.

\subsection{Performance Evaluation}
\label{sec:performance_evaluation}

\subsubsection{Single-Modal Recognition}
\label{sec:omni_single_modal}

Table~\ref{tab:omni_single_modal} reports single-modal recognition for all
nine inputs. This evaluation examines whether joint omni-modal training
retains discriminative identity descriptors for each individual modality.

Under BG, the appearance modalities are the most discriminative, with RGB
reaching $99.0\%$ Rank-1 accuracy, while the silhouette and geometric
modalities occupy an intermediate range and the event and radar
modalities remain weak. Under CL, IR provides the strongest result at
$55.7\%$, while LiDAR and depth retain comparatively useful geometric cues.
Notably, the unified RGB branch reaches $45.3\%$ Rank-1 accuracy under CL,
outperforming the RGB-specific DeepGaitV2-P3D baseline by $6.8$ percentage points.

The overall pattern is consistent with the task-specific study in
Section~\ref{sec:modality_evaluation}: joint training retains meaningful
modality-dependent strengths and failure modes within one model.

\subsubsection{Cross-Modal Recognition}
\label{sec:omni_cross_modal}
Table~\ref{tab:omni_cross_modal} reports all twelve directed retrieval
tasks. The Cross-Modal Avg. gives equal weight to the six modality pairs.

Among the six modality pairs, RGB and IR silhouettes achieve the
strongest compatibility, reaching $78.0\%$ bidirectional-average Rank-1
accuracy under BG. Retrieval between silhouettes and geometric
modalities is more difficult, while depth and LiDAR also remain
imperfectly aligned despite both encoding body geometry.
Compared with the pair-specific CL-Gait results in
Table~\ref{tab:eval_cross_modal}, OmniGait++ performs better on five of the
six modality pairs under both probe conditions, with RGB Sil.$\leftrightarrow$IR
Sil. as the only exception.

These results show that the jointly trained identity space enables direct
retrieval across heterogeneous modalities without pair-specific
fine-tuning. They also confirm that standalone discriminability and shared
physical content do not by themselves guarantee cross-modal compatibility.

\begin{table*}[!t]
	\centering
	\caption{Comparison with OmniGait under Protocol V2. Each cell is Rank-1 /
		mAP (\%). The common single-modal, cross-modal, and Two-Stream columns
		average the same nine modalities, six modality pairs, and eight fusion
		configurations, respectively. OmniGait does not support
		variable-cardinality fusion beyond pairwise anchor--auxiliary
		configurations.}
	\label{tab:comparison_omnigait}
	\small
	\renewcommand{\arraystretch}{1.0}
	\setlength{\tabcolsep}{2pt}
	\resizebox{0.99\textwidth}{!}{%
		\begin{tabular}{@{}l cc cc cc cc cc cc@{}}
			\toprule
			\multirow{2}{*}{\textbf{Method}} &
			\multicolumn{2}{c}{\textbf{Single-Modal}} &
			\multicolumn{2}{c}{\textbf{Cross-Modal}} &
			\multicolumn{2}{c}{\textbf{Two-Stream}} &
			\multicolumn{2}{c}{\textbf{Within-Group}} &
			\multicolumn{2}{c}{\textbf{Across-Group}} &
			\multicolumn{2}{c}{\textbf{Omni-Fusion}} \\
			\cmidrule(lr){2-3}\cmidrule(lr){4-5}\cmidrule(lr){6-7}
			\cmidrule(lr){8-9}\cmidrule(lr){10-11}\cmidrule(lr){12-13}
			& BG & CL & BG & CL & BG & CL & BG & CL & BG & CL & BG & CL \\
			\midrule
			OmniGait & 57.2/58.0 & 16.8/19.3 &
			50.6/51.7 & 11.1/13.4 &
			89.6/89.3 & 30.1/33.2 &
			--/-- & --/-- & --/-- & --/-- &
			--/-- & --/-- \\
			OmniGait++ & \textbf{65.5/66.0} & \textbf{28.8/31.3} &
			\textbf{57.0/57.9} & \textbf{19.7/22.3} &
			\textbf{95.5/95.3} & \textbf{56.1/58.7} &
			\textbf{97.6/97.4} & \textbf{67.3/69.5} &
			\textbf{97.1/97.0} & \textbf{62.2/64.5} &
			\textbf{99.5/99.5} & \textbf{80.8/82.5} \\
			\bottomrule
	\end{tabular}}
	\vspace{-2ex}
\end{table*}

\subsubsection{Multi-Modal Recognition}
\label{sec:omni_multi_modal}
Table~\ref{tab:omni_multi_modal} reports all 19 fusion configurations evaluated
using the same checkpoint and fusion module. For each of the four registry
groups, we additionally report the unweighted mean over its constituent
configurations for each metric and probe condition.

Every Two-Stream configuration surpasses the single-modal results of both its
RGB-silhouette anchor and corresponding auxiliary modality
under BG and CL. The gains are particularly
pronounced under clothing change: RGB Sil.+IR reaches $75.4\%$ Rank-1
accuracy, while RGB Sil.+Radar improves the anchor from $37.8\%$ to
$54.7\%$ despite the weak standalone performance of radar. Standalone
strength therefore does not fully determine a modality's value in fusion.
Compared with the separately trained MultiGait++ models in
Table~\ref{tab:eval_multimodal}, OmniGait++ performs better when RGB or IR
appearance is used as the auxiliary modality, but generally trails the
task-specific models for the remaining auxiliaries.

The higher-cardinality results further show that both composition and
cardinality matter. Within-group appearance/contour and geometry fusion are
more effective than motion/structure fusion, while the across-group settings
provide different gains according to their constituent cues. In the nested
omni-fusion sequence, adding RGB and IR appearance to Omni-5 improves CL
Rank-1 by $7.1$ percentage points, whereas further adding event and pose in
Omni-9 contributes only $0.5$ percentage points. This diminishing increment indicates
that auxiliary cues become partly redundant once appearance, contour, and
geometry are jointly available.

\subsubsection{Comparison with OmniGait}
\label{sec:comparison_with_omnigait}

Table~\ref{tab:comparison_omnigait} compares OmniGait++ with its preliminary
version, OmniGait, under the same Protocol V2. For their three shared operating
modes, the averages cover nine single-modal
settings, six cross-modal pairs, and eight Two-Stream configurations.

OmniGait++ consistently improves recognition in all three overlapping modes.
The gains are larger under clothing change, where single- and cross-modal
Rank-1 accuracy each increase by more than $8$ percentage points and
Two-Stream recognition rises from $30.1\%$ to $56.1\%$.
More importantly, OmniGait is restricted to fixed fusion between an RGB
silhouette anchor and one auxiliary modality, whereas OmniGait++ extends the
same checkpoint to within-group, across-group, and omni-fusion inputs of
different compositions and cardinalities. OmniGait++ is therefore not merely
a stronger pairwise model, but a transition toward unified omni-modal
recognition.

\subsection{Ablation Study}
\label{sec:ablation_study}

Table~\ref{tab:ablation_omnigait} evaluates the main design choices of
OmniGait++. Each block changes one factor from the full model while retaining
the remaining settings, and reports the same category-level averages used
above. Because one checkpoint must serve all operating modes, we select the
default configuration according to its consistency across tasks and probe
conditions rather than an isolated maximum.

\subsubsection{Ablation Study on Fusion Stage}
\label{sec:ablation_fusion_stage}

Here, Stage~$s$ denotes fusion immediately after private Stage~$s$, with the
remaining backbone stages forming the shared encoder. We evaluate fusion after
Stages~2--4. Fusion immediately after Stage~1 is not included: its
feature maps contain four times as many spatial locations as those of Stage~2,
and retaining dense spatial values for multiple variable-cardinality fusion
subsets incurs prohibitive GPU memory consumption under the same training
protocol.
Fusion at Stage~3 provides the strongest overall balance across the six task
groups. Earlier fusion at Stage~2 yields a small advantage on BG cross-modal
retrieval, but is less effective for most fusion settings, suggesting that the
features have not yet received sufficient modality-specific adaptation.
Conversely, delaying fusion to Stage~4 causes broad degradation, reducing
Omni-Fusion Rank-1 under CL by $8.6$ percentage points. At this boundary, no
shared backbone stage remains after aggregation to reconcile the heterogeneous
inputs. We therefore use Stage~3 to balance modality-specific adaptation with
shared post-fusion identity encoding.

\subsubsection{Ablation Study on Modality-Aware Normalization}
\label{sec:ablation_modality_aware_normalization}

Having established the fusion boundary, we next examine how features from
different modalities are normalized along the shared pathway. Modality-aware
normalization in the shared encoder is particularly important
for heterogeneous identity alignment. Removing it reduces cross-modal BG
Rank-1 from $57.0\%$ to $43.2\%$ and also degrades all fusion groups. Removing
only the modality-aware BNNeck has little effect on single- and cross-modal
recognition, with slight improvements in a few cross-modal entries, but
consistently weakens fusion, especially at higher cardinalities. Removing both
normalization components produces the deepest cross-modal degradation. These
results show that applying modality-aware normalization to the shared encoder
and BNNeck better accommodates heterogeneous modality statistics and improves
consistency across the unified operating modes.

\subsubsection{Ablation Study on Part-Aware Temporal Fusion}
\label{sec:ablation_part_aware_fusion}

With the encoder interface fixed, we turn to the internal fusion design,
beginning with its spatial granularity. Collapsing the feature height into one
region ($R{=}1$) consistently harms fusion, including a drop of $5.5$
percentage points in Omni-Fusion Rank-1 under CL. This
confirms that global temporal aggregation can obscure spatially localized gait
cues. Increasing the number of regions to 4, 8, or 16 substantially narrows
the gap, but the improvements are not monotonic: overly fine partitioning
provides no consistent benefit. We adopt $R{=}8$ because it gives the most
stable performance across the fusion groups, balancing local correspondence
with sufficient information in each region.

\subsubsection{Ablation Study on Anchor Exclusion}
\label{sec:ablation_anchor_exclusion}

After determining the regional granularity, we examine the roles assigned to
the anchor and auxiliary modalities in attention. Including the
RGB-silhouette anchor in the key--value set degrades every
fusion group under CL; for example, Across-Group Rank-1 decreases from
$62.2\%$ to $56.7\%$. Keeping the anchor query-only prevents its self-response
from competing with auxiliary evidence. The attention module consequently
estimates a correction from the auxiliary modalities, while the
residual path preserves the anchor features. This separation is
especially beneficial when the auxiliary set contains multiple heterogeneous
modalities.

\begin{table*}[!t]
	\centering
	\caption{Ablation of OmniGait++ under Protocol V2. Each cell is Rank-1/mAP
		(\%). Full denotes the default setting (fusion at Stage~3, modality-aware
		normalization in both the shared encoder and BNNeck, $R{=}8$ horizontal fusion regions,
		and the anchor excluded from the key--value set). In (d), Anchor
		$\notin$ KV is the default query-only setting.}
	\label{tab:ablation_omnigait}
	\footnotesize
	\renewcommand{\arraystretch}{1.0}
	\setlength{\tabcolsep}{2.5pt}
	\resizebox{0.99\textwidth}{!}{%
	\begin{tabular}{@{}l cc cc cc cc cc cc@{}}
		\toprule
		\multirow{2}{*}{\textbf{Setting}} &
		\multicolumn{2}{c}{\textbf{Single-Modal}} &
		\multicolumn{2}{c}{\textbf{Cross-Modal}} &
		\multicolumn{2}{c}{\textbf{Two-Stream}} &
		\multicolumn{2}{c}{\textbf{Within-Group}} &
		\multicolumn{2}{c}{\textbf{Across-Group}} &
		\multicolumn{2}{c}{\textbf{Omni-Fusion}} \\
		\cmidrule(lr){2-3}\cmidrule(lr){4-5}\cmidrule(lr){6-7}
		\cmidrule(lr){8-9}\cmidrule(lr){10-11}\cmidrule(lr){12-13}
		& BG & CL & BG & CL & BG & CL & BG & CL & BG & CL & BG & CL \\
		\midrule
		\multicolumn{13}{l}{\textit{(a) Fusion stage (private-to-shared boundary)}} \\
		\midrule
		Stage 2 & 63.8/64.2 & 28.7/31.3 &
		\textbf{57.9/58.7} & 19.5/22.1 &
		94.9/94.7 & 54.3/56.9 &
		96.9/96.7 & 63.5/65.7 &
		96.2/96.0 & 58.9/61.4 &
		99.2/99.1 & 77.2/79.1 \\
		Stage 3 & \textbf{65.5/66.0} & \textbf{28.8/31.3} &
		57.0/57.9 & \textbf{19.7/22.3} &
		\textbf{95.5/95.3} & \textbf{56.1/58.7} &
		\textbf{97.6/97.4} & \textbf{67.3/69.5} &
		\textbf{97.1/97.0} & \textbf{62.2/64.5} &
		\textbf{99.5/99.5} & \textbf{80.8/82.5} \\
		Stage 4 & 63.9/64.4 & 24.6/27.1 &
		47.6/49.1 & 16.0/18.6 &
		94.7/94.5 & 52.4/55.0 &
		97.1/97.0 & 62.8/65.3 &
		96.5/96.4 & 57.9/60.4 &
		99.3/99.3 & 72.2/74.5 \\
		\midrule
		\multicolumn{13}{l}{\textit{(b) Modality-aware normalization (abbreviated as MN)}} \\
		\midrule
		Full & \textbf{65.5/66.0} & \textbf{28.8/31.3} &
		57.0/57.9 & 19.7/22.3 &
		\textbf{95.5/95.3} & \textbf{56.1/58.7} &
		\textbf{97.6/97.4} & \textbf{67.3/69.5} &
		\textbf{97.1/97.0} & \textbf{62.2/64.5} &
		\textbf{99.5/99.5} & \textbf{80.8/82.5} \\
		w/o Enc. MN & 63.8/64.4 & 26.5/29.2 &
		43.2/45.1 & 13.9/16.4 &
		94.4/94.3 & 52.5/55.2 &
		96.7/96.6 & 63.8/66.1 &
		95.9/95.8 & 57.5/60.1 &
		99.1/99.1 & 76.6/78.6 \\
		w/o Neck MN & 65.2/65.8 & 28.7/31.3 &
		\textbf{57.6/58.4} & \textbf{19.9/22.5} &
		95.4/95.2 & 55.6/58.3 &
		97.5/97.4 & 66.8/69.1 &
		97.1/96.9 & 61.3/63.9 &
		99.4/99.4 & 78.9/80.7 \\
		w/o MN & 64.1/64.7 & 27.3/29.9 &
		41.7/43.5 & 13.0/15.5 &
		94.7/94.5 & 53.0/55.9 &
		96.9/96.8 & 64.3/66.6 &
		96.2/96.1 & 58.1/60.7 &
		99.3/99.2 & 75.6/77.7 \\
		\midrule
		\multicolumn{13}{l}{\textit{(c) Part-aware temporal fusion ($R$: number of horizontal fusion regions)}} \\
		\midrule
		$R{=}1$ & 64.8/65.4 & 27.8/30.4 &
		57.0/57.9 & 19.0/21.6 &
		93.5/93.1 & 53.0/55.4 &
		96.0/95.7 & 62.2/64.3 &
		95.3/95.0 & 58.8/61.0 &
		98.9/98.8 & 75.3/77.1 \\
		$R{=}4$ & 65.4/65.9 & 27.8/30.5 &
		\textbf{57.7/58.6} & 19.3/22.0 &
		95.3/95.1 & 55.6/58.3 &
		97.4/97.3 & 66.6/68.9 &
		97.0/96.9 & 61.6/64.0 &
		99.4/99.4 & 80.0/81.8 \\
		$R{=}8$ & \textbf{65.5/66.0} & \textbf{28.8/31.3} &
		57.0/57.9 & 19.7/22.3 &
		\textbf{95.5/95.3} & \textbf{56.1/58.7} &
		\textbf{97.6/97.4} & \textbf{67.3/69.5} &
		\textbf{97.1/97.0} & \textbf{62.2/64.5} &
		\textbf{99.5/99.5} & \textbf{80.8/82.5} \\
		$R{=}16$ & 65.2/65.8 & 28.0/30.6 &
		57.6/58.4 & \textbf{19.8/22.4} &
		95.4/95.2 & 55.0/57.6 &
		97.4/97.3 & 66.0/68.2 &
		97.0/96.8 & 60.9/63.3 &
		99.3/99.3 & 79.3/81.1 \\
		\midrule
		\multicolumn{13}{l}{\textit{(d) Anchor exclusion (KV: key--value set)}} \\
		\midrule
		Anchor $\notin$ KV & \textbf{65.5/66.0} & \textbf{28.8/31.3} &
		\textbf{57.0/57.9} & \textbf{19.7/22.3} &
		\textbf{95.5/95.3} & \textbf{56.1/58.7} &
		\textbf{97.6/97.4} & \textbf{67.3/69.5} &
		\textbf{97.1/97.0} & \textbf{62.2/64.5} &
		\textbf{99.5/99.5} & \textbf{80.8/82.5} \\
		Anchor $\in$ KV & 65.0/65.5 & 27.9/30.5 &
		56.3/57.2 & 19.1/21.8 &
		95.4/95.3 & 51.9/54.7 &
		97.2/97.1 & 63.1/65.6 &
		96.7/96.6 & 56.7/59.4 &
		99.4/99.3 & 77.4/79.4 \\
		\bottomrule
	\end{tabular}}
\vspace{-2ex}
\end{table*}

\section{Scope and Future Directions}
\label{sec:limitations_and_future_work}

\textbf{Broader Acquisition Settings.}
MMGait's matched multi-sensor recordings support controlled comparisons across
sensors, modalities, views, and walking conditions. By holding participant
identity and recording context consistent across sensing streams, this design
reduces confounding factors when analyzing modality-specific discriminability
and complementarity. Future data collections can complement this controlled
setting with greater environmental and device diversity, enabling systematic
evaluation of transfer across capture domains and deployment conditions.

\textbf{Native Structured Modalities.}
OmniGait++ establishes a common learning and fusion interface for nine
image-like modalities spanning all five sensing streams. Using a common
backbone family makes it possible to study modality adaptation, identity
alignment, and variable-cardinality fusion under a consistent architecture.
The modular private front-end also provides a natural insertion point for
native point-cloud tokenizers for LiDAR and radar, as well as a
coordinate-based tokenizer for 3D pose. These additions would extend the same
shared identity and fusion space to all twelve MMGait modalities while
preserving their native structures.

\textbf{Flexible Modality Availability.}
The 19-configuration registry systematically varies auxiliary composition and
cardinality around a common RGB-silhouette anchor, while the set-based fusion
design remains independent of fixed auxiliary branches. In particular, the
shared attention module accepts auxiliary sets with different sizes and
compositions without introducing configuration-specific parameters. This
structure provides a basis for studying broader availability patterns,
including alternative anchors and asymmetric probe--gallery modality subsets.
Incorporating modality-quality estimates could further allow the model to
adjust the contribution of noisy or degraded observations.

\textbf{Responsible Use.}
MMGait's provisions for informed consent, pseudonymization, controlled access,
and participant rights are detailed in Section~\ref{sec:mmgait_ethics}. These
measures provide a governance framework for research involving sensitive gait
data. Future dataset extensions and system deployments should preserve these
protections while documenting data access, model evaluation, and intended-use
boundaries.

Together, these directions extend MMGait and OmniGait++ toward broader, more
flexible, and responsible omni-modal gait recognition.

\section{Conclusion}
\label{sec:conclusion}

This work presents MMGait as a common experimental basis for heterogeneous gait
recognition. MMGait combines matched observations from five sensing streams and
twelve modalities with an identity-disjoint split and an impostor-augmented
protocol, enabling single-modal, cross-modal, and multi-modal recognition to be
compared under consistent evaluation. The results show that standalone
discriminability, cross-modal compatibility, and fusion complementarity are
distinct properties: standalone strength does not necessarily predict
cross-modal alignment or value in fusion.

To unify recognition across heterogeneous modality inputs and operating modes,
we developed OmniGait++. Its private-to-shared architecture preserves
modality-dependent cues while learning comparable identity descriptors, and
its anchor-guided temporal fusion accommodates modality subsets with different
sizes and compositions. Experiments demonstrate that one jointly trained
checkpoint supports this unified coverage while remaining competitive with
separately trained task-specific models.

Overall, MMGait provides the empirical foundation for systematically studying
heterogeneous gait information, while OmniGait++ offers a scalable framework
for learning from it. Together, they establish a common basis for advancing
omni-modal gait recognition.


\onecolumn
\setcounter{section}{0}
\setcounter{subsection}{0}
\setcounter{table}{0}
\setcounter{figure}{0}
\setcounter{equation}{0}
\renewcommand{\thesection}{S\arabic{section}}
\renewcommand{\thetable}{S\arabic{table}}
\renewcommand{\thefigure}{S\arabic{figure}}
\renewcommand{\theequation}{S\arabic{equation}}
\section*{Supplementary Material}

\section{Computational Cost}
\label{sec:supp_computational_cost}

Table~\ref{tab:computational_cost} compares the model size and path-wise
inference complexity of the task-specific baselines, OmniGait, and
OmniGait++. Total Params denotes the parameter count of one independently
trained checkpoint for a task-specific baseline and that of the complete
unified checkpoint for OmniGait and OmniGait++; it is therefore fixed across
their operating settings. Active Params denotes the union of the parameters
required by a specified operating path, with shared parameters counted once
and repeated execution reflected in FLOPs. Both parameter measures exclude the
BNNeck and identity classifiers.

FLOPs are measured before the BNNeck using batch size one. Each selected
modality provides a 30-frame input under the default
preprocessing setting. Gallery matching is excluded. For each configuration,
we count one forward pass through every distinct descriptor path required by
the setting: one modality path for single-modal inference, two modality paths
without fusion for cross-modal inference, and one fused path over the selected
modalities for multi-modal inference.

All quantities are measured per configuration and then averaged with equal
weight within each operating setting. Single-modal Avg., Cross-modal Avg., and Two-Stream Avg.
cover nine modalities, six unordered modality pairs, and eight fusion
configurations, respectively. Task-specific parameter counts are averaged per
checkpoint rather than summed over the expert collection, whereas Omni-5,
Omni-7, and Omni-9 are reported directly. \emph{For Omni-9, the inactive single-modal
normalization branches make Active Params slightly lower than Total Params.}

\begin{table}[!ht]
	\centering
	\caption{Model size and path-wise inference cost. Parameter counts and FLOPs
		exclude the BNNeck and identity classifiers; FLOPs also exclude gallery
		matching. Each modality provides 30 frames with batch size one. ``Avg.'' is
		an equal-weight average over the applicable configurations.}
	\label{tab:computational_cost}
	\setlength{\tabcolsep}{6pt}
	\renewcommand{\arraystretch}{1.3}
	\resizebox{0.6\textwidth}{!}{%
	\begin{tabular}{@{}llccc@{}}
		\toprule
		\textbf{Method} & \textbf{Operating Setting} &
		\textbf{\tabincell{c}{Total\\Params (M)}} & \textbf{\tabincell{c}{Active\\Params (M)}} &
		\textbf{FLOPs (G)} \\
		\midrule
		DeepGaitV2-P3D & Single-modal Avg. & 8.03 & 8.03 & 55.85 \\
		CL-Gait        & Cross-modal Avg.  & 10.20 & 10.20 & 114.78 \\
		MultiGait++    & Two-Stream Avg.   & 11.09 & 11.09 & 105.58 \\
		\midrule
		\multirow{3}{*}{OmniGait}
		& Single-modal Avg. & 9.14 & 7.23 & 73.16 \\
		& Cross-modal Avg.  & 9.14 & 7.46 & 146.25 \\
		& Two-Stream Avg.   & 9.14 & 7.51 & 105.45 \\
		\midrule
		\multirow{6}{*}{OmniGait++}
		& Single-modal Avg. & 24.12 & 8.27  & 82.22 \\
		& Cross-modal Avg.  & 24.12 & 9.98  & 164.37 \\
		& Two-Stream Avg.   & 24.12 & 12.15 & 135.24 \\
		& Omni-5            & 24.12 & 17.28 & 280.95 \\
		& Omni-7            & 24.12 & 20.70 & 378.16 \\
		& Omni-9            & 24.12 & 24.11 & 475.30 \\
		\bottomrule
	\end{tabular}}
\end{table}

The two parameter measures reveal complementary aspects of scalability.
Although the full OmniGait++ descriptor model is larger than an individual
task-specific descriptor model, covering the shared evaluation settings
requires nine single-modal experts, six pair-specific cross-modal experts, and eight Two-Stream experts.
Using the rounded per-model averages in Table~\ref{tab:computational_cost},
their aggregate size is approximately
$9(8.03)+6(10.20)+8(11.09)=222.19$M parameters, compared with 24.12M for
the full OmniGait++ descriptor model. Relative to storing all 23 task-specific
descriptor models, the unified model therefore reduces descriptor-model
storage by approximately $89\%$ while additionally supporting
variable-cardinality fusion.

At the level of an active path, OmniGait++ remains close to the corresponding
task-specific models.
Relative to OmniGait, the additional capacity and computation
accompany broader modality coverage and the consistent recognition gains in
Table~\ref{tab:comparison_omnigait}. Moreover, expanding from Omni-5 to Omni-7
and from Omni-7 to Omni-9 adds 3.42M and 3.41M Active Params and 97.21G and
97.14G FLOPs, respectively, indicating predictable growth as pairs of
modalities are added. Overall, OmniGait++ trades additional per-path
computation for substantially lower collection-level storage and broader
coverage within one checkpoint.

\section{Additional MMGait Visualizations}
\label{sec:supp_mmgait_visualization}

\begin{figure}[!ht]
	\centering
	\includegraphics[width=\linewidth]{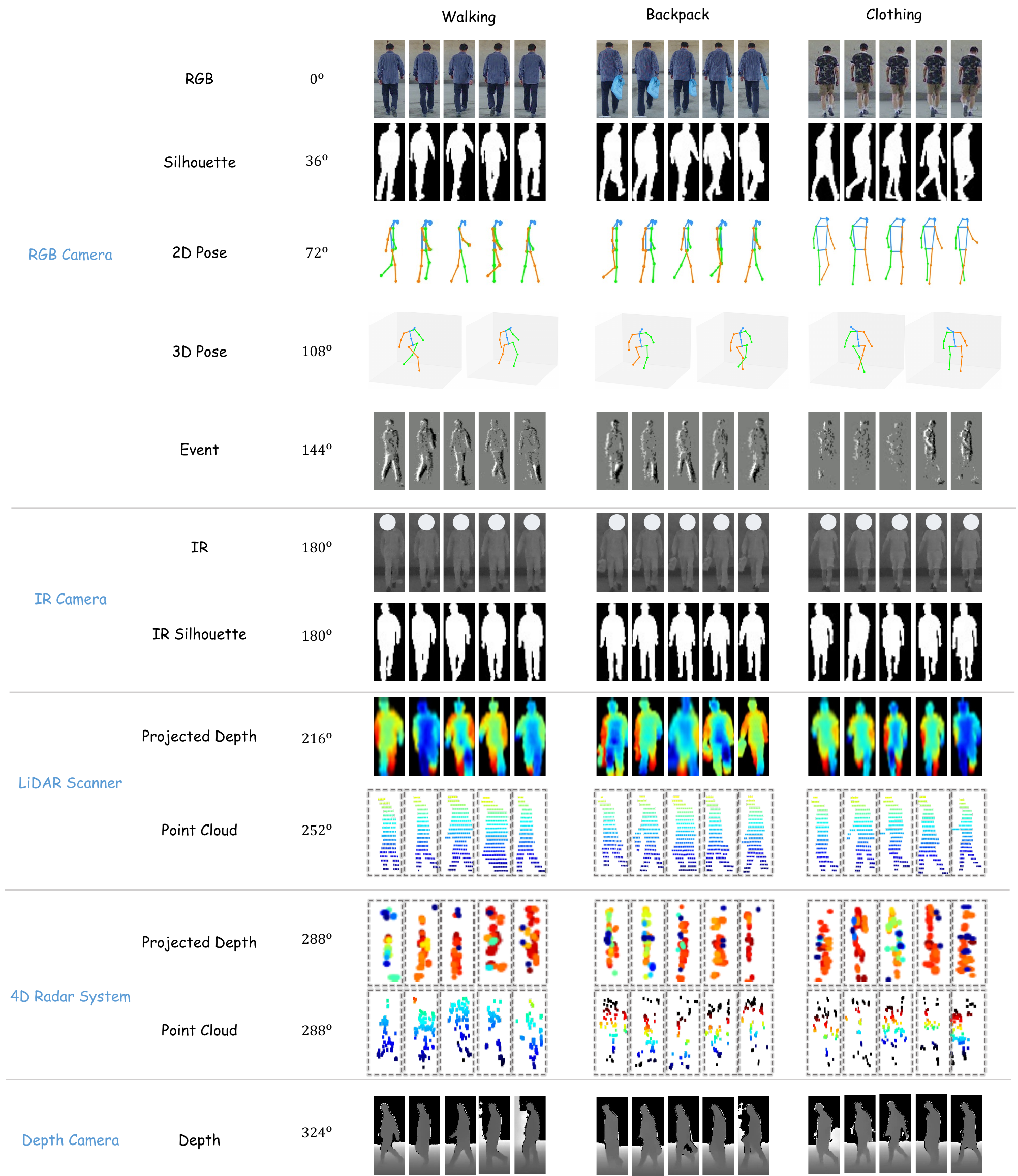}
	\caption{Representative MMGait sequences for one participant. Rows cover
		the twelve raw or derived gait modalities at the annotated walking
		directions, while columns show normal walking, backpack carrying, and
		clothing change.}
	\label{fig:mmgait_visualization_a}
\end{figure}

\begin{figure}[!ht]
	\centering
	\includegraphics[width=\linewidth]{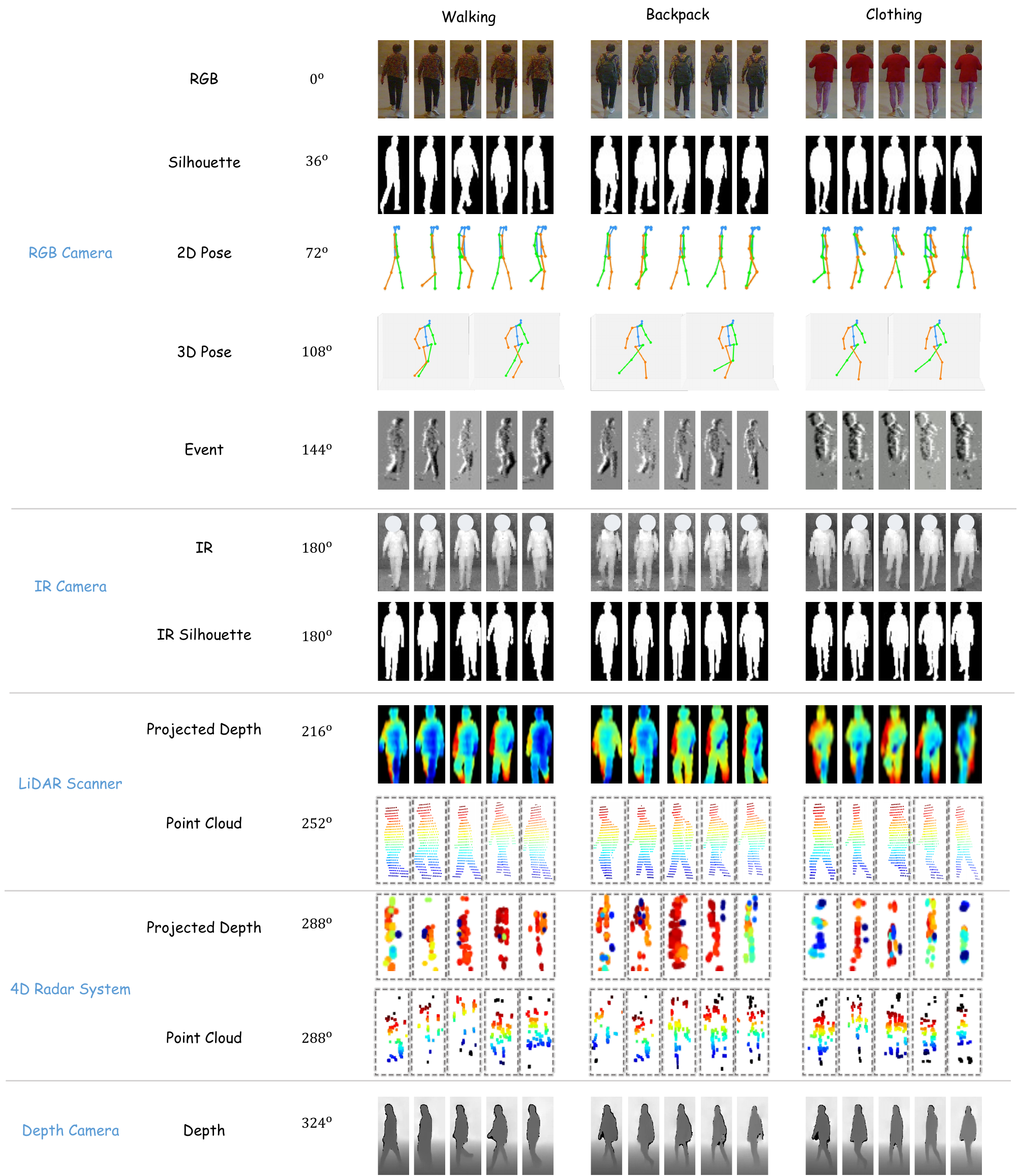}
	\caption{Additional representative MMGait sequences from another
		participant, showing the twelve raw or derived gait modalities at
		the annotated walking directions under normal walking, backpack
		carrying, and clothing change.}
	\label{fig:mmgait_visualization_b}
\end{figure}

Figures~\ref{fig:mmgait_visualization_a}
and~\ref{fig:mmgait_visualization_b} present processed examples from two
participants using the same layout. Rows group the raw and derived
modalities by sensing stream and indicate their walking directions,
whereas columns compare normal walking, backpack carrying, and clothing
change.

\end{document}